\documentclass[12pt]{article}
\usepackage{amsmath}

\DeclareMathOperator*{\argmin}{arg\,min}
\usepackage{amsfonts}
\usepackage{times}
\usepackage{graphicx}
\usepackage{color}

\usepackage{float}

\usepackage{supertabular}

\usepackage[inkscapelatex=false]{svg}

\usepackage{array, booktabs, makecell}

\usepackage{supertabular}

\usepackage{soul}
\usepackage{color}

\usepackage{tikz, pgfplots}
\usetikzlibrary{positioning}
\pgfplotsset{compat=1.18}
\usepackage{nccmath, mathtools}
\usepackage{comment}

\usepackage[authoryear]{natbib}
\usepackage{rotating}
\usepackage{bbm}
\usepackage{latexsym}

\usepackage{natbib}
\setcitestyle{authoryear,open={(},close={)}} 

\DeclareGraphicsExtensions{.pdf}

\newcommand{\captionfonts}{\normalsize}

\makeatletter  
\long\def\@makecaption#1#2{%
  \vskip\abovecaptionskip
  \sbox\@tempboxa{{\captionfonts #1: #2}}%
  \ifdim \wd\@tempboxa >\hsize
    {\captionfonts #1: #2\par}
  \else
    \hbox to\hsize{\hfil\box\@tempboxa\hfil}%
  \fi
  \vskip\belowcaptionskip}
\makeatother   

\begin{document}
\hspace{13.9cm}1

\ \vspace{20mm}\\

{\LARGE Contrastive Learning to Fine-tune Feature Extraction Models for the Visual Cortex}

\ \\
{\bf \large Alex Mulrooney} \hfill {\bf \large Austin J.\ Brockmeier} \\
{University of Delaware} \hfill  {University of Delaware}

\thispagestyle{empty}
\markboth{}{NC instructions}

\begin{center} {\bf Abstract} \end{center}
Predicting the neural response to natural images in the visual cortex requires extracting relevant features from the images and relating those feature to the observed responses. In this work, we optimize the feature extraction in order to maximize the information shared between the image features and the neural response across voxels in a given region of interest (ROI) extracted from the BOLD signal measured by fMRI. We adapt contrastive learning (CL) to fine-tune a convolutional neural network, which was pretrained for image classification, such that a mapping of a given image's features are more similar to the corresponding fMRI response than to the responses to other images. We exploit the recently released Natural Scenes Dataset \citep{allen2022massive} as organized for the Algonauts Project \citep{gifford2023algonauts}, which contains the high-resolution fMRI responses of eight subjects to tens of thousands of naturalistic images. We show that CL fine-tuning creates feature extraction models that enable higher encoding accuracy in early visual ROIs as compared to both the pretrained network and a baseline approach that uses a regression loss at the output of the network to tune it for fMRI response encoding. We investigate inter-subject transfer of the CL fine-tuned models, including subjects from another, lower-resolution dataset~\citep{gong2023large}. We also pool subjects for fine-tuning to further improve the encoding performance. Finally, we examine the performance of the fine-tuned models on common image classification tasks, explore the landscape of ROI-specific models by applying dimensionality reduction on the Bhattacharya dissimilarity matrix created using the predictions on those tasks \citep{mao2024training}, and investigate lateralization of the processing for early visual ROIs using salience maps of the classifiers built on the CL-tuned models. 

\section{Introduction}
Recent advances in computing power, large accessible datasets, and machine learning (especially deep learning) methods allow for new computational models of the brain. Modern approaches enable more powerful decoding of stimuli from neural responses~\citep{scotti2023reconstructing} and prediction of neural responses from stimuli via encoding models~\citep{Tang2023}. In visual neuroscience, the recently released Natural Scenes Dataset \citep{allen2022massive} provides a large dataset of brain responses to colored natural imagery for eight subjects measured by 7T fMRI BOLD responses (1.8mm-wide voxels, 1.6s sampling interval). The high resolution and high number of trials (close to 10,000 per subject) coupled with the complexity and breadth of imagery allows for more fine-grained modeling using deep learning. To this end, we propose a novel neural encoding approach that adapts a contrastive learning (CL) loss function~\citep{chen2020simple} to fine-tune a pretrained convolutional neural network (CNN) to produce feature representations of stimulus images that better align with the voxel-wise neural activity in specific regions of interest (ROIs) in the visual cortex. We fine tune a different instance of the CNN for each ROI, with different versions for each subject along with a pooled subject model. 

Our method differs from previous approaches to fMRI encoding \citep{kay2008identifying, mitchell2008predicting} that use predetermined feature maps to build linear models for individual voxels. We still use linear models for predicting the activity of individual voxels from the fine-tuned image features, but we fine-tune CNNs to act as ROI-specific non-linear feature extraction models for each subject and ROI, with ROIs spanning the entire visual cortex. Our approach is motivated by the fact that different ROIs in the visual cortex encode different visual characteristics. Once fine-tuned using CL, the set of ROI- and subject-specific CNNs are used for encoding and are analyzed in terms of the accuracy and similarity of their predictions on downstream image classification tasks, which can be visualized as a landscape of models~\citep{mao2024training}. In particular, we fine-tune a pretrained CNN, specifically AlexNet trained on ImageNet  \citep{krizhevsky2012imagenet}, a relatively simple deep learning architecture, optimizing both the convolutional and fully-connected layers. But the CL approach and the model landscape are general approaches and can be applied to other CNNs and feature extraction models beyond CNNs, such as a Vision Transformer~\citep{dosovitskiy2020image}, and can also be applied to other stimulus modalities such as speech and sound.   

For comparison, we use as a baseline the image features obtained from the pretrained, but untuned, AlexNet, and as another baseline we use an approach that fine-tunes the CNN using a regression loss at its output to predict the voxel activations in the ROI. With CL-based tuning, we note consistent improvements in encoding performance for early ROIs compared to the untuned AlexNet baseline. This result holds for another lower-resolution fMRI dataset, the Natural Object Dataset (NOD)~\citep{gong2023large}.  In comparison, regression-based fine-tuning leads to overfitting.  \\
Our contributions are as follows:
\begin{itemize}
\item We introduce a novel neural encoding approach that uses CL to fine-tune a CNN for ROI-specific feature extraction.
\item We show that models trained with CL enable significant improvements in encoding performance in terms of average correlation between predicted response and actual response on held-out test images versus the untuned baseline across ROIs early in the visual processing stream (V1, V2, V3, and hV4). 
\item The CL-tuned feature-extraction models for one subject and ROI are mostly transferable to the same ROI of other subjects (when using subsequent subject-specific linear encoding) with small losses in the percentage of improved voxels versus the untuned baseline. Some subjects are even better predicted by the CL-tuned models from other subjects for the same ROI, possibly due to higher noise levels in their own responses limiting the fine-tuning.
\item We use 9 subjects from the Natural Object Dataset~\citep{ge2024on}, which use atlas-based ROIs, to validate the CL-tuned feature extraction models of early visual ROIs, manually delineated, for one subject in the Natural Scenes Dataset~\citep{allen2022massive}. A linear model using the fine-tuned model's features achieves  a higher encoding performance of fMRI responses than untuned AlexNet in 8 of the 9 subjects tested, with an average improvement of the average correlation of 7\%. This indicates that ROI-specific CL fine-tuned models extract image features that are generally more predictive of the brain response in early visual areas, and thus more similar, than models solely trained for image classification.
\item After CL-based ROI-specific fined-tuning, we examine the CNNs' predictions and  accuracy on three image classification tasks (ImageNet ILSVRC2012~\citep{deng2009imagenet}, Caltech256~\citep{ griffin2007caltech}, and Places365 \citep{zhou2017places}) with freshly trained classification heads, in comparison to the untuned model. The lower accuracy obtained by the CL models meaningfully vary by ROI, with larger decreases for early ROIs. 
\item We visualize landscapes of the untuned and ROI-specific feature extraction models by performing dimensionality reduction on the Bhattacharyya dissimilarity created from pairwise comparisons of the downstream task predictions~\citep{mao2024training}. We note meaningful arrangement of early versus later visual ROIs and divergence between the early visual ROIs in the left and right hemispheres.
\item We highlight the potential for in-silico experimentation with ROI-specific fine-tuned models by examining the visual salience of predictions of the downstream image classifiers using Grad-CAM~\citep{selvaraju2017grad} to investigate lateralization in the models for early visual ROIs. 
\end{itemize}

\section{Related Work}
\subsection{Linear encoding models for fMRI}
Encoding models for the visual cortex that attempt to predict the neural response to an image often rely on two-stage modeling~\citep{la2022feature}. First, a predetermined nonlinear transformation is used to extract relevant features from stimulus images. Second, for each voxel, a linear model  is trained to predict the voxel response as a linear function of the image features. Typically, ordinary least squares or ridge regression~\citep{hoerl1970ridge} ($\ell_2$-regularized mean-squared error minimization) are used as the feature covariance matrix inversion is shared across all voxels.  The nonlinear feature mapping may vary based on what ROI the voxel is in, and can be informed by the hypothesized role of the ROI in question. The work by \citet{kay2008identifying} used a receptive-field model based on Gabor wavelets to predict activity of voxels in early visual regions, namely V1, V2, and V3. Semantic labels that describe basic object and action categories for an image have been used as features for encoding ROIs higher in the visual processing stream \citep{mitchell2008predicting}. Along similar lines, the work by \citet{lescroart2015fourier} compared linear encoding models using Fourier power characteristics, subjective distance, and object categories as different possible feature representations and found that while the Fourier characteristics were more predictive of voxels in V1, the object categories were more predictive of voxels in ROIs higher in the visual processing stream like RSC, OPA, and PPA (an ROI associated with encoding information about the spatial content of scenes \citep{kravitz2011real}). Multiple feature spaces can be used together through group ridge regression~\citep{la2022feature,Li2022}.

\subsection{Neural network-based encoding models}
More recently, representations from convolutional neural networks (CNNs) have been used as the basis for linear encoding models \citep{eickenberg2017seeing, gucclu2015deep}.  The work by \citet{eickenberg2017seeing} extracted the activations from one layer of a CNN, applied spatial subsampling for dimensionality reduction, then fit a ridge regression model using the features to predict the activations of each voxel. The work by \citet{gucclu2015deep} used a similar approach but without the intermediate dimensionality reduction. The works by \citet{eickenberg2017seeing} and \citet{gucclu2015deep} showed that the hierarchy of layers of CNNs trained on object recognition tasks had correspondences with and could be used to predict voxels throughout the human visual processing hierarchy. Specifically, they showed that the early convolutional layers of CNNs tended to be more predictive of voxels in early-visual regions, and the later layers in CNNs tended to be more predictive of ROIs higher in the visual processing stream.

Recently, more modern approaches that use various neural network architectures have been proposed for fMRI encoding in the visual cortex. These approaches allow for more flexibility in the learned feature representations of stimulus images. The work by \citet{zhang2019visual} used the pretrained AlexNet as the feature extraction model for predicting voxel activations in early visual regions. The work by \citet{han2019variational} trained a variational auto-encoder that can be used for both encoding brain responses and decoding viewed images, although they found lower encoding accuracies with this approach when compared to a CNN model. \citet{Dwivedi2021} used different deep neural networks with the same architecture trained for different vision tasks and found correspondence between the task and ROI, for example, edge detection for V1 and semantic tasks for later layers. At the same time as the writing of this paper, the work by \citet{st2023brain} showed that neural networks trained from scratch to predict neural activity in V1-V4 using various types of hierarchies can outperform pretrained AlexNet. Our work differs in the choice of CL for fine-tuning, which has not been seen in any previous work on encoding models. However, for the purpose of fMRI decoding, a recent paper used CL to learn embeddings of fMRI responses in the image space of a pretrained CLIP model that could be used for decoding and reconstruction of the visual stimuli \citep{scotti2023reconstructing}.

\subsection{Contrastive learning}
Contrastive learning (CL)~\citep{gutmann2010noise} is a framework that has been adopted for self-supervised learning \citep{ma2018noise,purushwalkam2020demystifying} and mutual information estimation~\citep{poole2019,mcallester2020formal}. The goal is that non-linear embeddings for one~\citep{chen2020simple} or more~\citep{radford2021learning} input spaces based on knowledge of positive pairings of input examples that share information and should be similar after embedding as contrasted to negative pairings of independently sampled examples. The positive and negative pairs may be constructed in various ways. In the SimCLR~\citep{chen2020simple} formulation,  CL is used for self-supervised learning on image datasets without labels. Positive pairs are created by associating two modified views of the same image, created by data augmentations like rotation, cropping, and color distortion. The augmentations are chosen such that they clearly encode the same information for downstream image classification tasks. The goal is to learn an embedding that ignores the variance due to augmentation and preserves the shared information between views. CL creates representations useful for downstream tasks~\citep{tosh2021contrastive}, which has been theoretically justified in the recent work by \citet{ge2024on}. In our case, the two views stem from the same image, but consist of the original stimulus images and the corresponding brain response. Therefore, a positive pair in our case is an image and its corresponding neural response (beta coefficients extracted from the fMRI BOLD response), and a negative pair is an image and a fMRI response for some other non-matching image. This is akin to pairing of an image with a caption, as in the CLIP model~\citep{radford2021learning}, which has been previously used for neural decoding of images from fMRI~\citep{ozcelik2023natural}.

\section{Methodology}

\subsection{Dataset}
The dataset used for the modeling and majority of our analysis is the Natural Scenes Dataset \citep{allen2022massive}, which consists of natural image stimuli and the average beta coefficients across multiple trials of a generalized linear model for the BOLD response captured by high-resolution 7T fMRI. Each of the eight subjects viewed approximately 9,000–10,000 distinct images across 3 trials (approximately 30,000 trials in total across numerous sessions). While beta coefficients were fit using voxel-specific hemodynamic response functions to each trial, our analysis uses the average across the 3 trials as provided~\citep{allen2022massive}. The  images used as stimuli are from the Common Objects in Context (COCO) database \citep{lin2014microsoft}. Data is available for voxels in the visual cortex that were responsive to the visual stimuli, with approximately 20,000 total voxels per subject across both left and right hemispheres. A wide array of ROIs are available, which were manually segmented using functional properties. Most ROIs have both a left hemisphere (LH) and right hemisphere (RH) group of voxels, and we treat each hemispheric region as its own ROI. The data was organized as part of the Algonauts Challenge, an open competition where participants attempted to predict visual cortex responses to complex natural images \citep{gifford2023algonauts}. We split the dataset (image and average response pairs) for each subject such that 85\% of the data is used for training and validation and the remaining 15\% is used for testing, with the testing set consistent across all ROIs for a given subject (but not across subjects, since different subjects viewed different images during the trials). The size of the test sets for subjects 1 through 8 are 1477, 1477, 1363, 1317, 1477, 1363, 1477, and 1317 images, respectively.

\subsection{Encoding model overview} \label{sec: model overview}
In our proposed approach, for each subject and ROI we fine-tune the parameters of an instance of a pretrained AlexNet CNN to act as the feature extraction model.  This results in 280 feature extraction models, where each model is specific to a single subject and hemisphere-specific ROI. 

For both our proposed approach and baselines we fit neural encoding models for each subject's ROIs using three stages: first, feature extraction is performed by passing the stimulus image to the pretrained or fine-tuned CNN, the activations at a particular layer are treated as features; second, as early CNN layers have high-dimensional activations, the number of dimensions is reduced via principal component analysis (PCA); and third, a linear encoding model is fit from these features to each voxel using ridge regression ($\ell_2$-normalized regression coefficients). 

To quantify the accuracy of the prediction, we evaluate the average correlation between the predicted responses for the validation data and the true responses and average across all voxels in the ROI, as described in Section~\ref{sec: accuracy metric}.

\subsection{Base and Baseline Feature Extraction Models} \label{sec: baseline models}
As a baseline feature extraction model we consider the AlexNet CNN~\citep{krizhevsky2012imagenet} pretrained on ImageNet~\citep{ILSVRC15}. All encoding models are specific to a particular subject- and hemisphere-specific ROI, but the base model (with or without fine-tuning) serves as a backbone for all image feature models. 

To match the expected input for AlexNet, the stimulus images are all resized to 224-by-224 pixels  and their means and standard deviations are normalized so that the means across the 3 colors channels are 0.485, 0.456, 0.406, and the standard deviations across the color channels are 0.229, 0.224, and 0.225 (a standard preprocessing step for using AlexNet pretrained on ImageNet). 

\subsubsection{Pretrained AlexNet Model}  \label{sec: best layer selection}
We consider using the different layers of AlexNet (either after one of the five convolutional layers or after one of the 3 fully connected layers) as shown in Table~\ref{tab:Alexnet layer dims} for feature extraction.
\begin{table}[H]
    \centering
    \begin{tabular}{@{}lrrrrrrrr@{}}
        \toprule
                \textbf{Layer} &Conv-1&Conv-2& Conv-3& Conv-4&  Conv-5& FC-1& FC-2&FC-3 \\
        \textbf{Dimension} &
         46,656 &
        32,448 &
        64,896 &
        43,264  &
       9216  &
        4096  &
         4096 &
        1000 \\ 
        \bottomrule
    \end{tabular}
    \caption{Dimensions of the output at 8 different AlexNet layers, where the spatially organized convolutional layers (conv) have been vectorized. The final layers are fully connected (FC), with the final output being 1000 dimensions.}
    \label{tab:Alexnet layer dims}
\end{table}
The pretrained AlexNet model acts as the starting point for all functions  performing image feature extraction. Tapping into the AlexNet model at layer $l\in \{1,\ldots,L\}$ with $L=8$, we can select different amounts of processing of the original image for different ROIs.

Before the linear encoding model is fit, for each layer $l\in\{1,\ldots,L\}$, we apply PCA to reduce the dimensionality of the representation to be constant for all layers, and take the first $d=1000$ principal components of the training images' representations for a subject (as training images differ by subject). PCA for dimension reduction of the activations is applied \textbf{after} fine-tuning.

For each ROI, we select the best representation layer for subsequent linear encoding from the pretrained AlexNet model by evaluating the cross-validated performance of an $\ell_2$-regularized linear encoding model fit for each voxel in the ROI on a representative subject using the dimension-reduced activations by evaluating the eight possible layers $l\in\{1,\ldots 8\}$. For simplicity, after fine-tuning, the same activation layer is used for fitting the linear encoding model, but a different layer may in fact provide better a fit.

\subsection{Mathematical Formulation of Neural Encoding Framework}
 Let $x\in \mathcal{I}$ represent a single image (for AlexNet, $\mathcal{I} \subset \mathbb{R}^{3\times 224 \times 224}$). For a given subject, let  $\mathbf{y}^r\in \mathbb{R}^{v^r}$ represent the corresponding $v^r$-dimensional cortical response in the $r$th ROI. The encoding model is a function $\mathit{Enc}^{r}: \mathcal{I}\rightarrow \mathbb{R}^{v^r}$ that predicts the $v^r$-dimensional cortical response as a composition of two models $\mathit{Enc}^r = \mathit{enc}^r\circ f^r$: the ROI-specific feature extraction model $f^r: \mathcal{I} \rightarrow \mathbb{R}^{d^r}$, where ${d^r}$ is the dimension of the feature extraction model, and the linear encoding model $\mathit{enc}^r: \mathbb{R}^{d^r} \rightarrow \mathbb{R}^{v^r}$. Let $\mathbf{a}^r=f^r(x)$ denote the activations at the output of $f^r$ and $\hat{\mathbf{y}}^r=\mathbf{b}^r+\mathbf{B}^r \mathbf{a}^r$ denote the prediction of the cortical response in the ROI to image $x$ given by the encoding model with linear coefficients $\mathbf{B}^r\in\mathbb{R}^{v^r\times d^r}$ and bias $\mathbf{b}^r\in\mathbb{R}^{v^r}$. Let $\mathcal{F}$ denote the family of {feature extraction} models. Ideally, the goal is to optimize the feature extraction model for the $r$th ROI that minimizes the sum of mean-squared error across voxels given an optimal regression for each voxel. This is stated as the following optimization problem:    
\begin{align} 
\label{feature extraction optimization}
\min_{f \in \mathcal{F}} \left \{  \min_{\substack{ \mathbf{b} \in \mathbb{R}^{v^r}, \\ \mathbf{B} \in\mathbb{R}^{v^r \times d^r} }}\mathbb{E}[\lVert \mathbf{y}^r - (\mathbf{b} +  \mathbf{B}f(x) )\lVert_2^2 ] = \sum_{k=1}^{v^r} \min_{\substack{ b\in \mathbb{R}, \\ \mathbf{w} \in \mathbb{R}^{d^r}}} \mathbb{E}[(y_k^r - (b+\mathbf{w}^\top f(x)))^2] \right \}.
\end{align}
In practice, limited training data $\{ ( x_i, \mathbf{y}_i^r ) \}_{i=1}^{n_{\mathrm{train}}} \subset \mathcal{I} \times \mathbb{R}^{v^r}$ is available to compute the expectation, and the capacity of $\mathcal{F}$, especially when it is a deep CNN, leads to overfitting, where the predictions on training data are accurate but the predictions on unseen data are inaccurate.  

\begin{figure}[htbp]
    \centering
    \includegraphics[width=\textwidth]{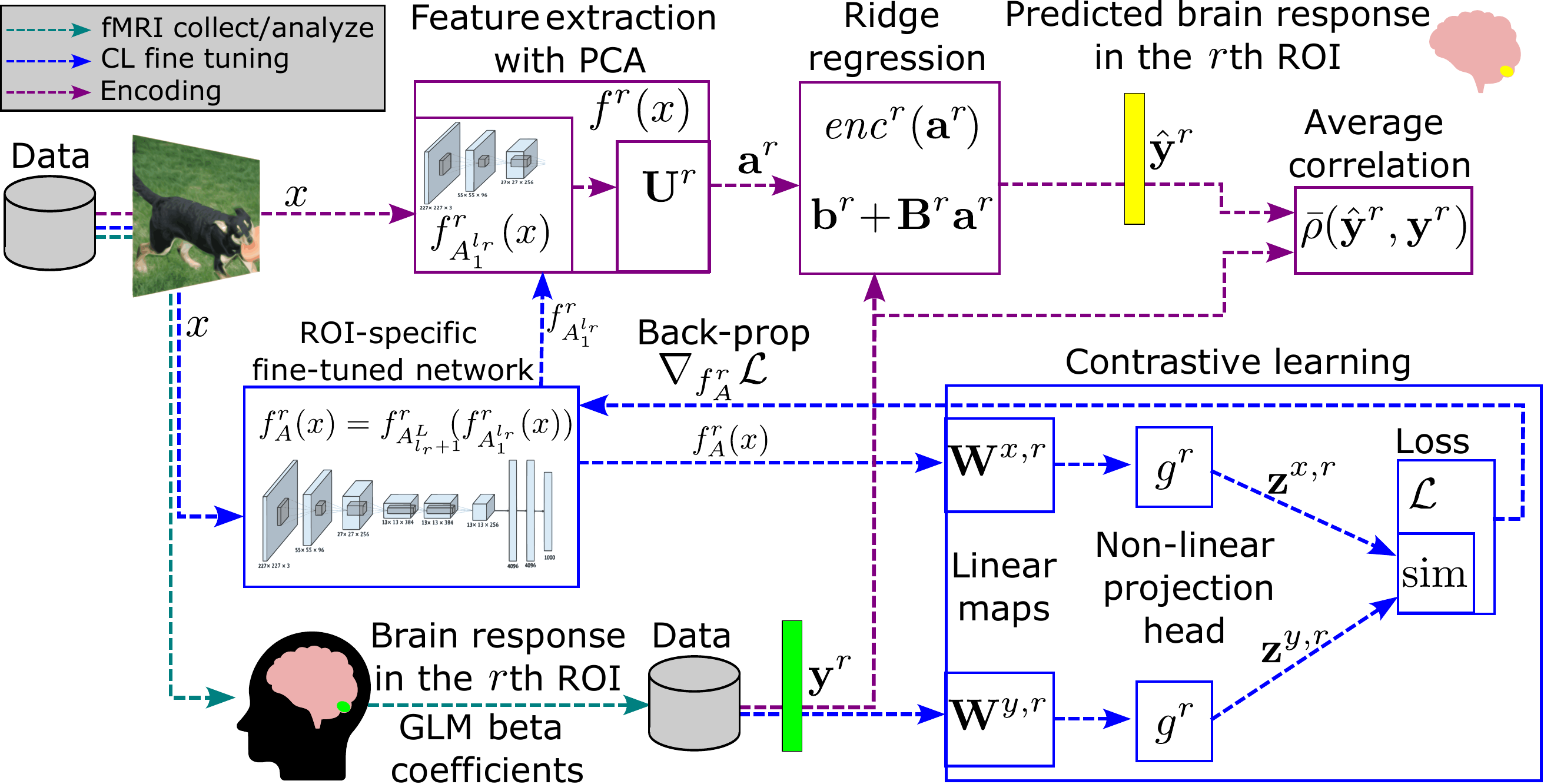}
    \caption{Overview of the proposed approach. (Green arrows) The brain response to natural images (GLM beta coefficients fit to the fMRI) are organized into visual ROIs. (Blue arrows) Constrastive learning (CL) is used to fine-tune the AlexNet CNN $f_A^r$ based on a loss $\mathcal{L}$ that maximizes the cosine similarity $\mathrm{sim}$ of processed image-response pairs $\mathbf{z}^{x,r},\mathbf{z}^{y,r}$, contrasted to random pairs $\mathbf{z}^{x',r},\mathbf{z}^{y,r}$ with $x'\neq x$ (not shown). $\mathbf{z^{x,r}}$ is obtained by passing the image $x$ through the AlexNet CNN $f_A^r$ and then a linear projection $\mathbf{W}^{x,r}$ and the shared non-linear projection head $g^r$. $\mathbf{z^{y,r}}$ is obtained by passing the fMRI response $\mathbf{y}^r$ through a linear projection $\mathbf{W}^{y,r}$ and then the shared non-linear projection head $g^r$. $f_A^r$ is updated by back-propagation using the gradient $\nabla_{f_A^r} \mathcal{L}$. (Purple arrows) An encoding model for the ROI uses the output from the pre-selected layer $l_r$ of the fine-tuned AlexNet, applies PCA, and then fits a $\ell_2$ penalized linear model via ridge regression. Performance of the encoding predictions are assessed on held-out images via the correlation coefficient averaged across an ROI's voxels $\bar{\rho}$.
    }
    \label{fig:main_figure}
\end{figure}

To improve generalization and prevent overfitting of the encoding models, we use a three stage approach. First, we select a particular layer $l\in\{1,\ldots,L\}$ of a pretrained CNN after the CNN has been fine-tuned with a contrastive learning loss function  to serve as the feature extraction model $f^r$; second, we apply principal component analysis (PCA) to reduce the dimension of the features; and third, we apply $\ell_2$-regularized linear regression to the reduced feature set to form the encoding model. The PCA and regularization both reduce the effect of overfitting, but the PCA is essential to reduce the computational demands of forming and inverting a high-dimensional covariance matrix. The overview of our approach is depicted in Figure~\ref{fig:main_figure}.

In the first stage of our proposed approach (and the regression-tuned baseline), an instance of the pretrained AlexNet model is fine-tuned for each ROI and individual.  Let $f_{A_{1}^l}:\mathcal{I}\rightarrow \mathbb{R}^{d_l}$ denote the processing performed by the pretrained AlexNet model at the $l$th layer and $f_{A_{l+1}^L}:\mathbb{R}^{d_l}\rightarrow \mathbb{R}^{d_L}$ denote the subsequent processing, such that $f_{A_1^L}= f_{A_{l+1}^L} \circ f_{A_1^l}$ is the full pretrained AlexNet model with output $d_L=1000$. Let $\mathcal{F}_A$ denote the set of all possible functions obtained by changing the AlexNet parameters. We denote the $r$th ROI fine-tuned model as $f^r_A=f^r_{A_1^L}=f^r_{A_{l_r+1}^L} 
\circ f^r_{A_1^{l_r}} \in \mathcal{F}_A$, where only the fine-tuned feature extraction model $f^r_{A_1^{l_r}}:\mathcal{I}\rightarrow \mathbb{R}^{d_{l_r}}$ {up } to the pre-selected layer $l_r$ is used for subsequent encoding. (Note that the ROI-specific layer {pre-selection} creates a {feature extraction} model with dimension $d_{l_r}=d^r$ {before PCA is applied}.) The reasons for this design choice are two-fold: first, the scale of the activations for a layer after fine-tuning will be more commensurate with the pretrained version given back-propagation through the subsequent layers, which is necessary for the {same} $\ell_2$-regularization {parameter} to be useful; second, the final layers of the network {will be} used for downstream image classification tasks to benchmark how the representation is changed based on fine-tuning. {However, it may be better to directly tune only a portion of the network based on an intermediate representation, or select a different layer after fine-tuning the entire network. Additionally, the choice of fixed regularization parameter for the entire ROI and after fine-tuning is likely suboptimal.}

Given an {ROI-tuned} feature extraction model consisting of  the fine-tuned AlexNet at the $l_r$th layer   $f^r_{A_1^{l_r}}:\mathcal{I}\rightarrow \mathbb{R}^{d^r}$, PCA is applied to reduce the dimension, yielding the  dimension-reduced  activation vector $\mathbf{a}^r= f^r(x)= \mathbf{U}^{r\top}f^r_{A_1^{l_r}}(x)$,  where $\mathbf{U}^r \in \mathbb{R}^{d^r \times d}, d\le d^r$ is the matrix of $d$ eigenvectors defining the principal components of the activations in the training set $\{ f^r_{A_1^{l_r}}(x_i)\}_{i=1}^{n_\mathrm{train}}$. Given this, a regularized linear model is fit with penalty $\alpha$ as
\begin{equation} \label{eq:ridge reg}
     \mathbf{b}^r,\mathbf{B}^r = \argmin_{ \mathbf{b}\in\mathbb{R}^{v^r}, \mathbf{B}\in\mathbb{R}^{v^r\times d}}
    \sum_{i=1}^{n_\mathrm{train}}  \lVert \mathbf{y}_i^r - (\mathbf{b}+\mathbf{B} \mathbf{a}_i^r) \rVert_2^2 + \alpha \lVert \mathbf{B} \rVert^F_2,
\end{equation}
where $\lVert \mathbf{B} \rVert^F_2$ denotes the squared Frobenius norm (sum of the squared weights) and $[\mathbf{a}_i^r]_{i=1}^{n_\mathrm{train}}  \in  \mathbb{R}^{d^r \times n_{\mathrm{train}}}$ are the dimension-reduced activations on the training data. The predicted neural response to image $x$ is then $\hat{\mathbf{y}}^r = \mathbf{b}^r + \mathbf{B}^r \mathbf{U}^{r\top}f^r_{A_1^{l_r}}(x)$.

\subsection{Accuracy metric} \label{sec: accuracy metric}
To assess the performance of a model, we compute the average correlation coefficient between the predicted response and the true response across all voxels in that ROI
\begin{equation}
\label{accuracy metric}
    \bar{\rho}(\hat{\mathbf{y}}^r, \mathbf{y}^r) = \frac{1}{v^r} \sum_{k=1}^{v^r}  \rho(\hat{y}_k^r, y_k^r),        
\end{equation}
where $\rho(\hat{y}_k^r, y_k^r)$ is the Pearson correlation coefficient between the predicted responses for the $k$th voxel of the $r$th ROI $\hat{y}_k^r=\mathit{Enc}_k^{r}(x)$ and the true responses $y_k^r$. Accuracies over the entire test set are computed as the average $\bar{\rho}(\hat{\mathbf{y}}^r, \mathbf{y}^r)$ over all test set images.

\subsubsection{Hyper-Parameter Selection: Dimension and Regularization}
For each ROI, we perform 5-fold cross validation using the training data to jointly select the best AlexNet output layer and the best penalty term $\alpha$ for the regularized linear regression using the log-spaced grid $[10^{-1}, 10^0, ..., 10^7]$ as a hyper-parameter combination. To reduce computation, the cross-validation is performed on every ROI {available for} subject 1, and we use the same hyper-parameter combination for the corresponding ROI in the other subjects. Since the ROIs vary somewhat over the subjects and not all of the ROIs are present in every subject (see Figure~\ref{fig:roi_subj_presence_table} in \ref{appA}), ROIs not present in subject 1 are taken from the lowest numbered subject that contains that ROI. 

Let $l_r, \alpha_r$ denote the hyper-parameter combination selected for $r$th ROI. We use this combination of the selected layer and penalty term chosen for the pretrained AlexNet models in subsequent linear models for both the regression-fine-tuned and the CL-fine-tuned models. However, it may be the case that regression- or CL-tuning of AlexNet models starting from different layers (or using different regularization) would perform better.

\subsubsection{Neural network regression model} \label{sec: neural network regression model}
Based on \eqref{feature extraction optimization}, {and to parallel the CL-approach, we create a fine-tuning baseline that tries to predict} the voxel activity in a given subject's $r$th ROI {from the final AlexNet activations of dimension $d=1000$}---note that after fine-tuning, intermediate output at layer $l_r$ will be used as the features for the actual linear encoding model. The {feature extraction model (and subsequent layers)} is optimized using the mean squared error between the predicted and actual fMRI response as the loss function, 
\begin{equation} \label{eq:reg model}
    f^r_A = \argmin_{f \in  \mathcal{F}_A }
   \min_{ \mathbf{b}_O \in\mathbb{R}^{v^r}, \mathbf{B}_O\in\mathbb{R}^{v^r\times d}} \sum_{i=1}^{n_\mathrm{train}}  \lVert \mathbf{y}_i^r - (\mathbf{b}_O+\mathbf{B}_Of(x_i)) \rVert_2^2, 
\end{equation}
where the parameters of both a linear mapping at the output and the entire AlexNet are optimized during training, with the latter initialized as the pretrained AlexNet CNN $f_{A_1^L}$. After fine-tuning, layer $l_r$ of the regression-tuned network serves as the feature extraction model $f^r = f_{A_{1}^{l_r}}^r:\mathcal{I}\rightarrow \mathbb{R}^{d_{l_r}}$, the ouptut is dimension-reduced via PCA and an encoding model is fit to predict voxel responses using the $\alpha_r$ penalty term for the $\ell_2$-regularization \eqref{eq:ridge reg}. While indirect, i.e., the encoding model built at the output of the fine-tuned CNN is not actually used,  using regression to fine-tune the image feature extraction model specific to a subject and ROI has the potential to outperform the baseline pretrained model, but not overfit as much as a direct encoding model as in \eqref{feature extraction optimization}. 

\subsection{Contrastive Learning for ROI-specific Feature Extraction} \label{sec: contrastive learning model}
The crucial technique in our proposed approach is the use of a contrastive loss function to fine-tune the ROI-specific feature extraction model. The goal is to learn representations of images that emphasize features that are relevant for a given ROI. Contrastive estimation uses machine learning to estimate log density functions~\citep{gutmann2010noise} based on discriminating `positive' true data from `negative' data created from noise or a source of randomness. In the context of paired random variables, the noise is independent observations, and estimates of the log-density of the conditional distribution have been shown to lower-bound mutual information~\citep{poole2019}. Maximizing this lower-bound provides an objective for optimizing the image feature {extraction} model.

Inspired by the SimCLR framework \citep{chen2020simple}, which optimizes a feature extraction network based on differently augmented pairs of the same original image as positive pairs and augmented pairs of different images as negative pairs, we propose a multimodal CL approach akin to the CLIP model~\citep{radford2021learning} that takes a  given image $x_i\in \mathcal{I}$ paired with its corresponding fMRI response $\mathbf{y}_i^r \in \mathcal{Y}^r \quad \forall r$ as a positive pair, and an image $x_i$ and a disparate fMRI response $\mathbf{y}_j^r$, $i \neq j$ as a negative pair.  The contrastive loss aligns the response in the $r$th ROI $\mathbf{y}^r$ to the output of $f_A^r$. After fine-tuning, the feature extraction model $f^r$ uses the features extracted at layer $l_r$ from $f_A^r$.

{Given a batch of $K$ paired vectors in the same space $\{(\mathbf{z}^x_i,\mathbf{z}^y_i)\}_{i=1}^K$, a contrastive loss inspired by SimCLR~\citep{chen2020simple} is
\begin{align}
\mathcal{L}(\{(\mathbf{z}^x_i,\mathbf{z}^y_i)\}_{i=1}^K)=
\frac{1}{K}\sum_{i=1}^K -\log \left( \frac{\exp( \frac{1}{\tau} \mathrm{sim}(\mathbf{z}^x_i,\mathbf{z}^y_i))}{ \sum_{j\neq i} \exp( \frac{1}{\tau} \mathrm{sim}(\mathbf{z}^x_i,\mathbf{z}^y_j) )}\right),
\end{align}
where} $\mathrm{sim}(\mathbf{z},\mathbf{z}')=\frac{\langle \mathbf{z},\mathbf{z}'\rangle}{\lVert  \mathbf{z}\rVert_2\lVert\mathbf{z}'  \rVert_2}$ is the cosine similarity and $\tau>0$ is a temperature parameter. {While slightly different due to the dropping of the positive pair $j\neq i$ in the denominator, this contrastive learning loss function can be related to a lower bound on mutual information~\citep{poole2019}, as discussed in \ref{sec:mi}.} 

To map the ROI-specific image features and the ROI response to a common space for cosine similarity we propose to use two linear mappings denoted by matrices $\mathbf{W}^{x,r}$ and $\mathbf{W}^{y,r}$ to map the fine-tuned activations $\mathbf{a}^r=f_A^r(x)$ and $\mathbf{y}^r$, respectively, to a shared vector space $\mathbb{R}^{d_h^r}$. {As in the SimCLR architecture~\citep{chen2020simple},} this is followed by a shared non-linear function, specifically, a multi-layer `perceptron' (MLP), referred to as a projection head $g^r : \mathbb{R}^{d_h^r} \rightarrow \mathbb{R}^{d_z^r}$ {(we note that CLIP~\citep{radford2021learning} does not use a projection head). The vectors in the shared space are then $\mathbf{z}^{x,r}=g^r(\mathbf{W}^{x,r} f_A^r(x))$ and $\mathbf{z}^{y,r}=g^r(\mathbf{W}^{y,r} \mathbf{y}^r)$.} The model architecture is summarized by Figure~\ref{fig:main_figure}. 

The optimization problem is
\begin{equation} 
\label{training objective}
f_A^r,\mathbf{W}^{x,r},\mathbf{W}^{y,r},g^r  = \argmin_{f \in \mathcal{F}_A, \mathbf{W}^x,\mathbf{W}^y,g } \mathbb{E}_{\pi}\left [  \mathcal{L}( \{ (g(\mathbf{W}^x f(x_{\pi(i)})) , g(\mathbf{W}^y\mathbf{y}^r_{\pi(i)}) ) \}_{i=1}^K  ) \right ],
\end{equation}
where the expectation is over mini-batches drawn from the training set defined by the mapping function $\pi: \{1,\ldots, K\} \rightarrow \{1,\ldots,n_\mathrm{train}\}$ created by sampling $K$ integers from $\{1,\ldots,n_\mathrm{train}\}$ without replacement. Based on the bound on mutual information (see \ref{sec:mi}), $K$ should be as large as possible \citep{poole2019,chen2020simple}, but is limited in practice since the computation of $\mathcal{L}$ has quadratic complexity with respect to the batch size.

Once the subject and ROI-specific {model $f_A^r$ is fine-tuned, the activations at the $l_r$th layer  are used as features for building the subsequent encoding model. Essentially the subsequent layers, which are also fine-tuned, are discarded for the purposes of encoding, as in the regression-tuned baseline. Dropping later fine-tuned layers has precedence in self-supervised learning~\citep{bordes2023guillotine}.  As in the other baselines, the activations at this layer} are projected onto the first $d=1000$ principal components {based on the training set} and $\ell_2$-regularization with penalty $\alpha_r$ is used to fit the encoding model.

\subsubsection{Implementation Details}
\label{sec:optimization hyper-params}

We set the dimensionality of the vector spaces based on $v_r\in\mathbb{N}$, the number of voxels {in the ROI}, with $d_h^r = \lfloor 0.8 v^r \rfloor$, $d_z^r = \lfloor 0.2 v^r\rfloor$, and the temperature as $\tau = 0.3$. {(In comparison, for SimCLR~\citep{chen2020simple}, $\tau\in\{0.1, 0.5, 1.0\}$, while for CLIP~\citep{radford2021learning}, $\tau$ is initialized to $0.07$ and limited to $100$ but updated in terms of the parameter $t=\log(\tau)$).} {The projection head consists} of three fully connected layers, with the first two layers each being followed by a batch normalization layer and then a ReLU activation function. The first two layers of the MLP projection head keep the dimension of the projection $h$ the same, and the final layer reduces the dimension to $d_z^r$.

We use Scikit-Learn to fit the regularized linear models \citep{pedregosa2011scikit} and PyTorch with CUDA to tune the neural networks \citep{paszke2019pytorch}. For all subjects and ROIs, the fine-tuning optimization uses batch sizes of $K=1024$, 30 epochs for CL and 75 epochs for regression, and an Adam optimizer with a learning rate of  $2.5\cdot10^{-5}$ for contrastive learning and $10^{-4}$ for regression, betas of 0.9 and 0.999, an epsilon of $10^{-8}$, and no weight decay. The number of training epochs {for each method} was selected by using a subset of 30 randomly selected ROIs from across the 8 subjects, splitting up the training data further into $85\%$ training and $15\%$ validation, and manually noting when the validation loss had converged for all the ROIs.  Learning rates were selected based on the validation loss for a few ROIs, but the grid search was not extensive. We fine-tune the models using T4, V100, and P100 GPUs, on which it takes roughly 2 minutes per epoch of training.

\subsection{Cross-subject transfer of fine-tuned feature extraction models} \label{sec: cross subject models}
We also investigate how well the learned stimulus space of a feature extraction model trained for a particular subject's ROI can be used to predict voxel activations for the same ROI in a different subject. To do this, we use the fine-tuned AlexNet activations at layer $l_r$ as features $f^{r,i}=f^{r,i}_{A_1^{l_r}}$ for subject $i$ for the desired ROI to extract image features from the images in subject $j$'s training set, and then compute the principal components, which differ since the images presented to each subject create different training sets,  and train an $\ell_2$-penalized linear model that predicts the voxel responses in subject $j$.

\subsection{Pooled models} \label{sec: pooled models}
For each ROI, we also train a pooled model using the data from all subjects which have that ROI to fine-tune a single shared AlexNet backbone. The architecture is identical to the one described for subject-specific models in Section~\ref{sec: contrastive learning model}, but with a separate linear mapping $\mathbf{W}^{y,r}_s$ for each ROI and subject $s$, which is required because of the varying voxel mappings and dimensionality for the same ROI across different subjects. To assess the impact of the dimensionality of the latent space $d_h$, for each ROI we train one pooled model with $d_h$ equal to the average voxel dimension of all subjects with that ROI (rounded to an integer), and another pooled model with $d_h$ equal to the maximum voxel dimension over all ROIs in all subjects. The same training methodology and optimization hyper-parameters as in Section~\ref{sec:optimization hyper-params} are used. Notably, one epoch of training involves a pass over the training images and corresponding fMRI responses for each subject, rather than just one subject, which increases the training set size. Combined with the cross-subject encoding models, the performance of the pooled models gives an indication of how transferable the features learned from fine-tuning on one subject are to other subjects.

\subsection{External Dataset Validation}
\label{sec:NOD-method}
A valid concern is whether the CL-tuned models are transferable beyond the Natural Scences Dataset with its high-resolution fMRI and its ROI definitions. In order to investigate whether the CL-tuned models are extracting images features that are predictive of the fMRI response in corresponding ROIs, we validate them for brain encoding on the Natural Object Dataset (NOD) \citep{gong2023large}, which uses the Human Connectome Project Multi-Modal Parcellation (HCP-MMP) for ROI definitions~\citep{glasser2016multi}. Notably, NOD uses a 3T scanner and also uses stimulus images drawn from ImageNet for some subjects. For our validation, we use the responses to the ImageNet images for subjects 1 through 9, each of which have recorded responses to 4000 unique ImageNet images (whereas the remaining 21 subjects in NOD only have responses to 1000 images each). For each subject, we split the 4000 responses into 3400 responses which are used for training and  hyper-parameter selection of the ridge regression penalty term $\alpha$, and 600 responses for testing the encoding performance. Because the fMRI responses are relatively noisy, we follow the encoding experiments \citep{gong2023large} and use only the 50 voxels in each ROI with the highest noise ceiling estimates. \\
Based on the results for NSD, we focus on the early visual ROIs V1-V4 from the ventral stream in both left and right hemisphere for each subject. For each ROI and subject, we use the same layer from AlexNet that was used to predict fMRI responses in the same ROI in the NSD experiment and perform 5-fold cross-validation over the training data to select the best ridge regression penalty term for using the untuned AlexNet activations to predict the fMRI responses in that subject's ROI. Then, using that penalty term, we fit $\ell_2$ penalized linear models for each subject and ROI using both untuned AlexNet, and the CL- and regression-tuned AlexNet models for the same ROI in the NSD experiment. This follows the same procedure as was used with NSD. For all subjects and ROIs, we use the CL- and regression-tuned models from subject 1 in NSD, since the models for subject 1 are most predictive of fMRI responses in early-visual cortex on average.

\subsection{Downstream classification tasks} \label{sec: downstream tasks}
To quantitatively analyze the CL-tuned feature extraction models, we use standard image classification tasks to build classifiers on top of the untuned and ROI fine-tuned models. We use three different image classification datasets:
ImageNet ILSVRC2012~\citep{deng2009imagenet}, which is the dataset used in the AlexNet pre-training, Caltech256~\citep{ griffin2007caltech}, and Places365 \citep{zhou2017places}. ILSVRC2012 contains 1000 different classes, Caltech256 has 255 object classes (plus 1 catch-all "clutter" class), and Places365 has 365 classes. The Caltech256 and Places356 datasets provide external validation of the fine-tuned models, with the former containing classes focusing on specific objects or figures like bears, helicopters, and microwaves, while the latter's categories are more general scenes like bedrooms, cafeterias, and staircases. We use the validation split of ILSVRC2012 containing 50,000 images, the entire Caltech256 dataset containing 30,607 images, and the validation portion of Places365, which has 36,500 images. For each, we randomly split the dataset and use 85$\%$ for training and 15$\%$ for testing. As features we use the penultimate layer's 4096-dimensional activations of the tuned or pretrained AlexNet CNN as feature representations of the images, fit a multinomial logistic regression model to predict the labels of the training data, and then evaluate the accuracy of the classifier in predicting the labels of the testing images. For ILSVRC2012, this allows us to probe how much fine-tuning (CL or regression) degrades the downstream performance as the model  focuses on features relevant for predicting the activity in a subject's ROI, possibly forgetting features useful for the classification task. For the Caltech256 and Places356 tasks, this accuracy indicates how well the features of the untuned and fine-tuned models can be used for datasets outside of the one that they were initially trained on. In summary, we train a classifier on top of the pretrained AlexNet CNN (replacing its own classification head), each subject-ROI-specific CL fined-tuned model, each pooled-subject, ROI-specific CL fine-tuned model (with both choices of embedding layer dimension) and each subject-ROI-specific regression-tuned model.

\subsection{Model Landscapes}
Given a set of classification models, \citet{mao2024training} propose an approach for visualizing the principal components of their pairwise dissimilarities based on the average similarity of their predictions across a test set. 

For $M$ models, the pairwise dissimilarity matrix  $\mathbf{D}\in[0,\infty)^{M\times M}$ is computed as the average of the negative logarithm of the Bhattacharyya similarity $\mathrm{B}$ of the class prediction probabilities across a test set of size $\\n_\mathrm{test}$, $D_{m,m'} = \frac{1}{n_\mathrm{test}} \sum_{i=1}^{n_\mathrm{test}}-\log( \mathrm{B}(\mathbf{p}^m(x_i),\mathbf{p}^{m'}(x_i))$, $m,m'\in \{1,\ldots,M\}$, where $\mathrm{B}(\mathbf{p},\mathbf{q})=\sum_{k=1}^C \sqrt{p_k q_k}$ and  $\mathbf{p}^m(x_i)$ represents the $m$th classification model's probabilities estimates across the $C$ classes for image $x_i$. Multidimensional scaling (MDS) is applied to derive a 2D embedding for the $m$th model with coordinates
$[U_{m1}\sqrt{|\Lambda_{11}|}, U_{m2}\sqrt{|\Lambda_{22}|}]$, where $U_{mp}$ is the $m$th entry of the eigenvector corresponding to the $p$th largest magnitude eigenvalue $\Lambda_{pp}$ of the matrix $-\frac{1}{2}\mathbf{H}\mathbf{D}\mathbf{H}=\mathbf{U}\mathbf{\Lambda}\mathbf{U}^\top \in \mathbb{R}^{M\times M}$ where $\mathbf{H}=\mathbf{I}_{M\times M}-\frac{1}{M}\mathbf{J}$ is the centering matrix and $\mathbf{J}$ is a matrix of all ones).

\subsection{Salience Maps for CL-tuned Models}
\label{sec:saliency methods}
As a simple example of the type of ``in-silico'' experiments that can be conducted with the CL-tuned models for different ROIs, we visualize the salient parts of an image in terms of the model's classification predictions using gradient-based class activation maps (Grad-CAM) \citep{selvaraju2017grad}. We use the CL-tuned model with the final layer weights from the trained linear classifier as described in Section \ref{sec: downstream tasks}, and then apply Grad-CAM.

\section{Results}
We divide the ROIs into two coherent groups~\citep{gifford2023algonauts}: early visual regions (V1, V2, V3, hV4) and higher visual regions (body-, face-, place-, or word-selective regions such as EBA, OFA, OPA, and OWFA, respectively). A full description of the ROI groupings is given in \ref{appA}. 

\subsection{Pretrained AlexNet Model}
\begin{figure}[H]
    \centering
    \includegraphics[scale=1.1]{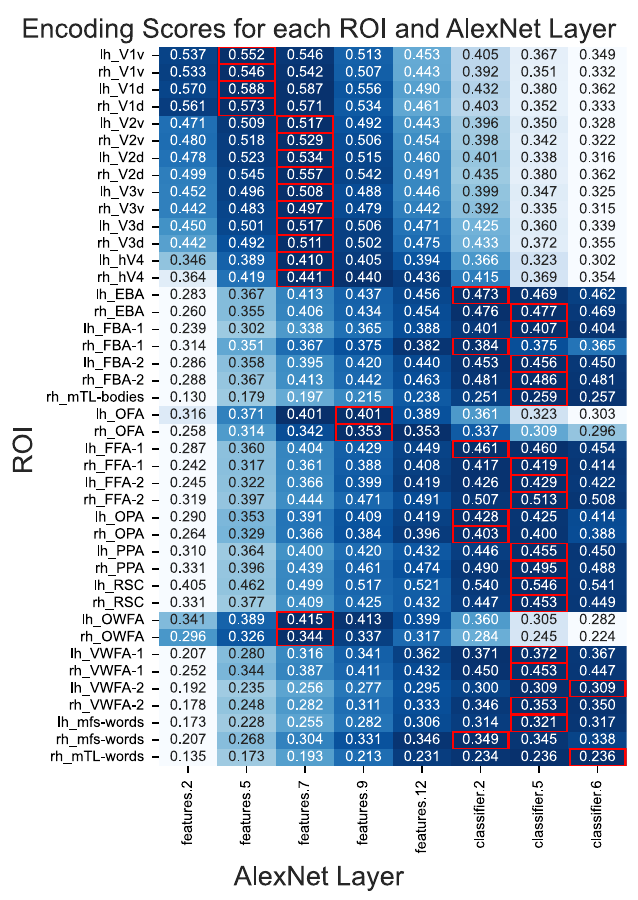}
    \caption{Matrix of encoding scores for each ROI and AlexNet layer, where the entries are the average encoding accuracy over 5-fold cross validation when a particular AlexNet layer's activations (after PCA for dimensionality reduction) are used to predict a particular ROI (using the best penalty term from cross validation for the regularization parameter for ridge regression).}
    \label{fig:untuned_layers_mat}
\end{figure}

As described in Section~\ref{sec: baseline models}, we perform 5-fold cross validation to jointly select the AlexNet output layer and the penalty term for the ridge regression for each ROI. We include the layers after each of the 5 convolutional layers and subsequent activation functions and the 3 fully connected layers for a search space of 8 possible output layers. Figure \ref{fig:untuned_layers_mat} shows the average cross validation encoding accuracy for each ROI and AlexNet output layer, with the best output layer for each ROI highlighted.
We find that all of the early visual ROIs (V1-hV4) are best predicted by either the second or the third of the five convolutional layers, and that 24 of the 28 higher visual ROIs (counting left- and right- hemisphere splits as different ROIs) are best predicted by the output of one of the two fully-connected layers. Our findings for the most predictive AlexNet layer agree with previous works that suggest that early-visual regions are more similar to the middle convolutional layers (Conv-2 or Conv-3) in CNNs, and ROIs higher in the visual processing stream (PPA, ventral) have higher similarity with the later fully-connected layers \citep{kriegeskorte2015deep, eickenberg2017seeing, gucclu2015deep}. 

\subsection{Subject-specific models}
\begin{figure}[htbp]
    \centering
    \begin{minipage}{0.24\textwidth}
        \centering
        \includegraphics[width=\textwidth]{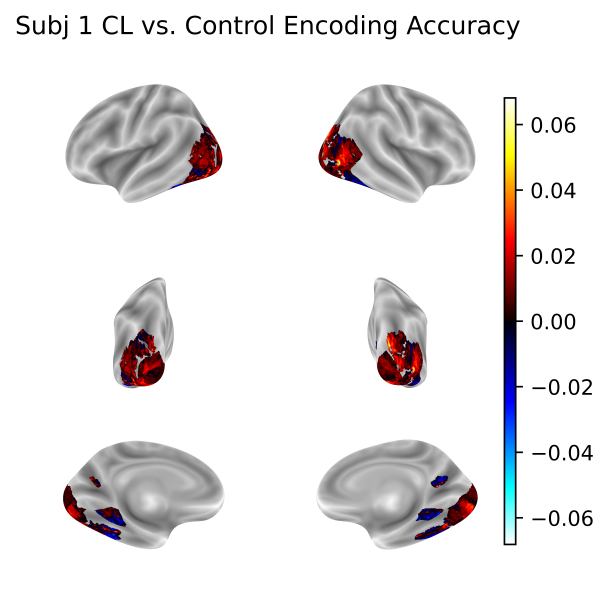}
    \end{minipage}
    \begin{minipage}{0.24\textwidth}
        \centering
        \includegraphics[width=\textwidth]{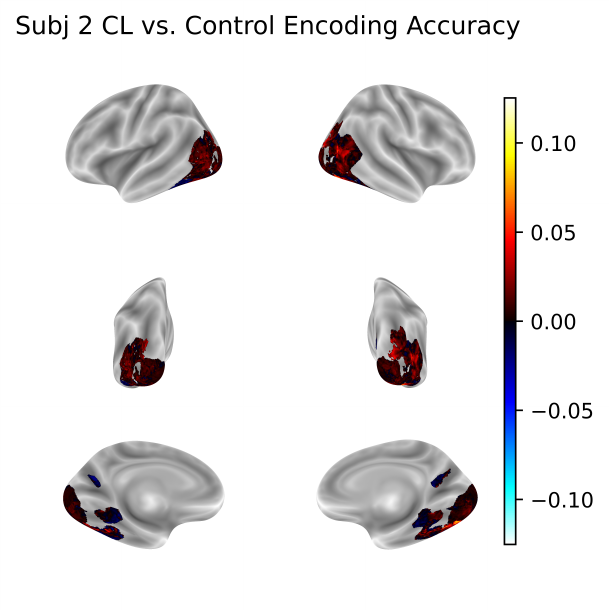}
    \end{minipage}
    \begin{minipage}{0.24\textwidth}
        \centering
        \includegraphics[width=\textwidth]{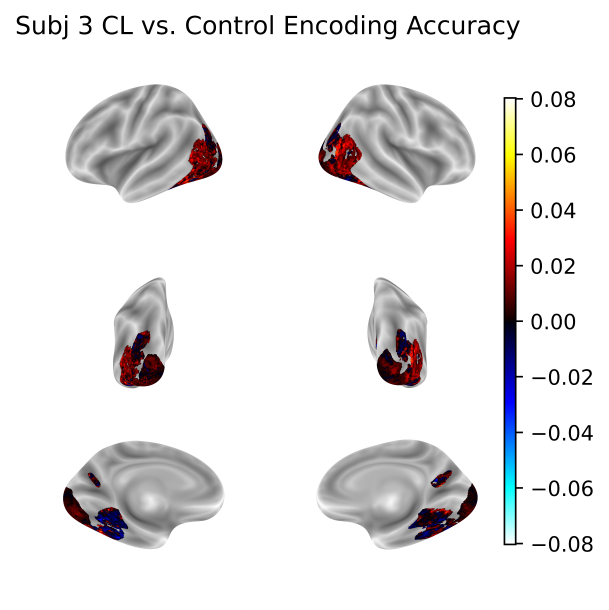}
    \end{minipage}
    \begin{minipage}{0.24\textwidth}
        \centering
        \includegraphics[width=\textwidth]{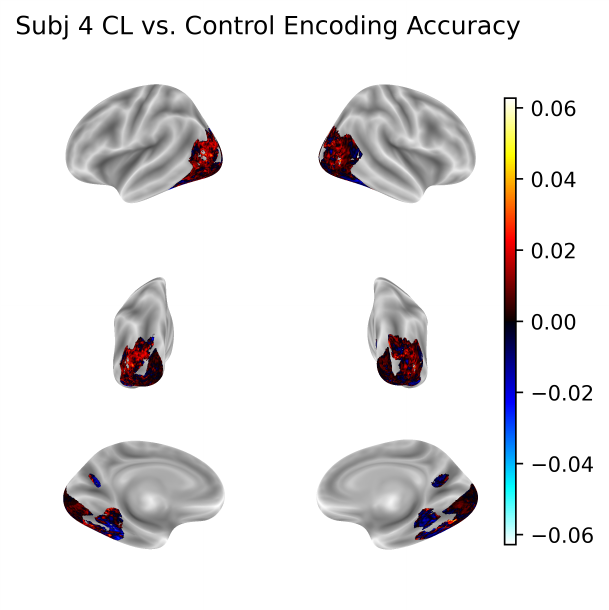}
    \end{minipage}
    \begin{minipage}{0.24\textwidth}
        \centering
        \includegraphics[width=\textwidth]{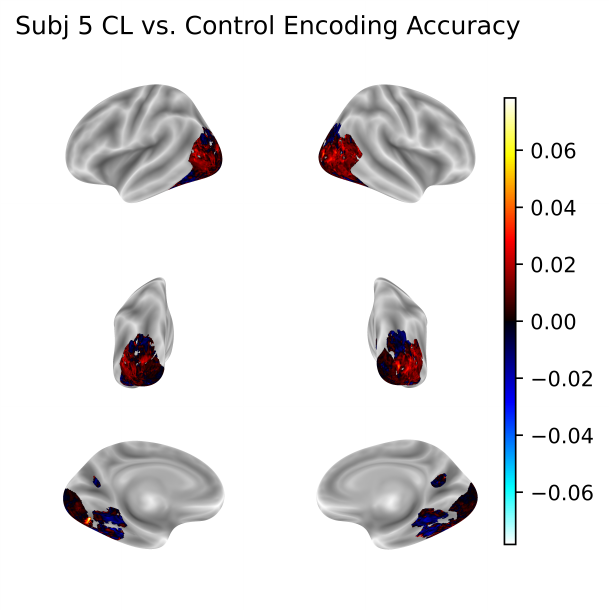}
    \end{minipage}
    \begin{minipage}{0.24\textwidth}
        \centering
        \includegraphics[width=\textwidth]{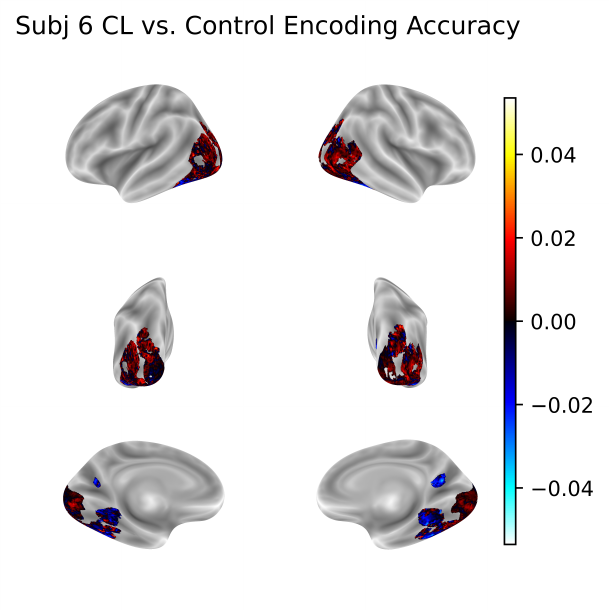}
    \end{minipage}
    \begin{minipage}{0.24\textwidth}
        \centering
        \includegraphics[width=\textwidth]{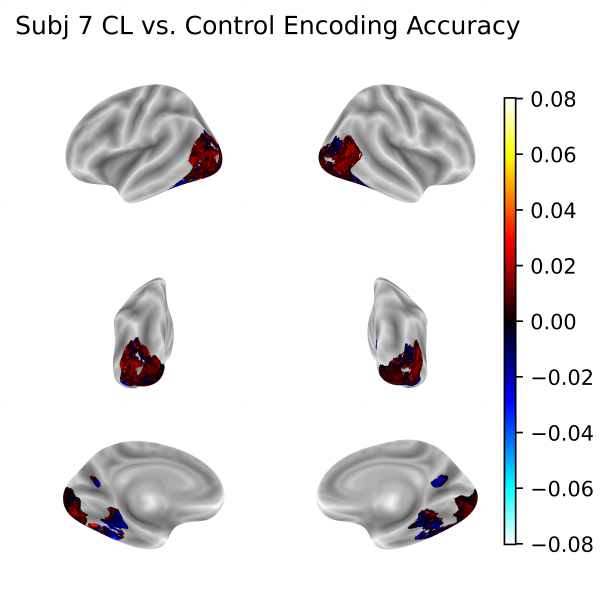}
    \end{minipage}
    \begin{minipage}{0.24\textwidth}
        \centering
        \includegraphics[width=\textwidth]{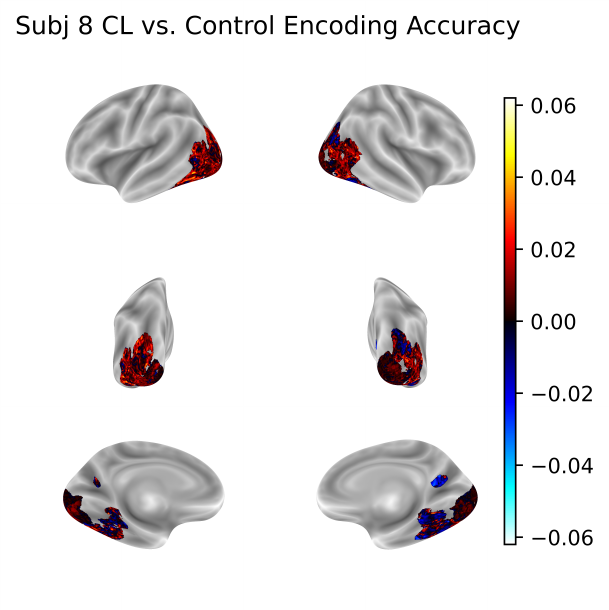}
    \end{minipage}
    
    \caption{Voxel-wise differences in encoding performance $\rho$  visualized on FreeSurfer template for each subject. `Hot' colors indicate improvements with contrastive learning (CL) based fine-tuning versus the pretrained AlexNet. }
    \label{fig:cortex_results}
\end{figure}

We first analyze the performance of the subject-specific models described in Section~\ref{sec: contrastive learning model} on the test set. For each ROI, we evaluate the average Pearson correlation coefficient $\bar{\rho}(\mathbf{y}^r, \hat{\mathbf{y}}^r)$  as described in Section~\ref{sec: accuracy metric} for both baseline models and the CL model, and also judge the performance of the CL model versus the baseline models by computing the percentage of voxels in the ROI that had a higher average test correlation than the baseline model in question across the test images. Subject-specific results for a given subject for each of these groups are calculated as the mean over all ROIs in the group. Table \ref{table:subject specific results} shows the results for the early and higher visual ROI groups. We also visualize the difference in encoding accuracy for each voxel across the cortex of each subject in Figure \ref{fig:cortex_results}, where we use the FreeSurfer template \citep{fischl2012freesurfer}.

\newpage
\vfill

\begin{table}[H]
    \begin{tabular}{@{}ccccccccccc@{}}
        \toprule
        \textbf{Sub.} &
        \multicolumn{2}{c}{\textbf{Ctrl. Avg.  $\bar{\rho}$}} &
        \multicolumn{2}{c}{\textbf{CL  Avg. $\bar{\rho}$ }} &
        \multicolumn{2}{c}{\textbf{Reg. Avg. $\bar{\rho}$}}  &
        \multicolumn{2}{c}{\textbf{CL $>$ Ctrl. }} & 
        \multicolumn{2}{c}{\textbf{ CL $>$ Reg.  }} \\ 
        \cmidrule(lr){2-3} \cmidrule(lr){4-5} \cmidrule(lr){6-7} \cmidrule(lr){8-9} \cmidrule(lr){10-11}  
        & \textbf{E} & \textbf{H} & \textbf{E} & \textbf{H} & \textbf{E} & \textbf{H} & \textbf{E} & \textbf{H} & \textbf{E} & \textbf{H} \\
        \midrule
        1 & {0.524} & 0.418 & {0.532} & 0.420 & {0.520} & 0.398 & {90.2}\% & 58.7\% & {94.3}\% & 90.3\% \\
        2 & 0.504 & 0.443 & 0.509 & 0.449 & 0.497 & 0.432 & 85.0\% & {59.7}\% & 92.7\% & 81.6\% \\
        3 & 0.420 & 0.373 & 0.424 & 0.374 & 0.411 & 0.349 & 79.7\% & 51.6\% & 93.7\% & 90.5\% \\
        4 & 0.402 & 0.362 & 0.403 & 0.361 & 0.397 & 0.335 & 65.8\% & 46.3\% & 83.8\% & 89.6\% \\
        5 & 0.448 & {0.476} & 0.450 & {0.477} & 0.443 & {0.463} & 66.2\% & 49.0\% & 88.5\% & 77.4\% \\
        6 & 0.404 & 0.343 & 0.406 & 0.340 & 0.396 & 0.312 & 68.1\% & 41.1\% & 89.4\% & 92.4\% \\
        7 & 0.402 & 0.363 & 0.406 & 0.361 & 0.395 & 0.334 & 75.7\% & 44.8\% & 90.3\% & {93.4}\% \\
        8 & 0.339 & 0.277 & 0.342 & 0.279 & 0.333 & 0.251 & 73.5\% & 54.4\% & 83.5\% & 90.0\% \\ \midrule
        All & 0.430 & 0.382 & 0.434 & 0.382 & 0.424 & 0.359 & 75.5\% & 50.7\% & 89.5\% & 88.2\% \\ \bottomrule
    \end{tabular}
    \caption{Subject-specific encoding performance for early (E) and higher (H) visual ROIs using  the untuned AlexNet (Ctrl.), see Section~\ref{sec: baseline models}; with CL-tuning (CL), see Section~\ref{sec: contrastive learning model}; and with regression-tuned (Reg.), see Section~\ref{sec: neural network regression model}. (Avg. $\bar{\rho}$) denotes the test set correlation $\bar{\rho}$, described in Section~\ref{sec: accuracy metric}, averaged across ROIs.  The last two columns show the percentage of voxels, averaged across ROIs in the group, for which encoding with the CL-tuned model was more predictive compared to either the untuned model or the regression-tuned model.}
    \label{table:subject specific results}
\end{table}

\clearpage

We find that the CL-based fine-tuning produces significantly higher encoding correlation coefficients than the baseline pretrained AlexNet approach in the early visual ROIs (corrected p-value of $0.0023$   and effect size 1.65 (Cohen's d) for a  paired, one-tailed t-test with Bonferroni correction for 2 tests, sample size of $8$ subjects, and significance threshold of $\alpha=0.1$) but not in the higher visual ROIs (corrected p-value of 0.48 and effect size of 0.27). We also examine the average percentage of improved voxels using CL versus the untuned baseline. 
For the early visual ROIs, the average percentage of improved voxels using CL versus the untuned AlexNet model across subjects is $75.5\%$. 
In the higher visual ROIs, the average percentage of improved voxels using CL versus the untuned AlexNet across subjects is $50.7\%$. which is at a chance level.

\subsection{Cross-subject transfer of fine-tuned feature extraction models}
\begin{figure}
    \centering
    \includegraphics[scale=0.66]{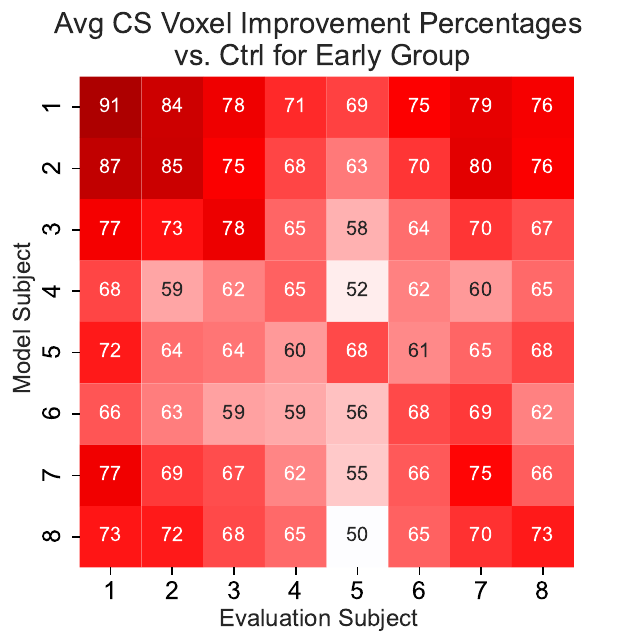}
    \includegraphics[scale=0.66]{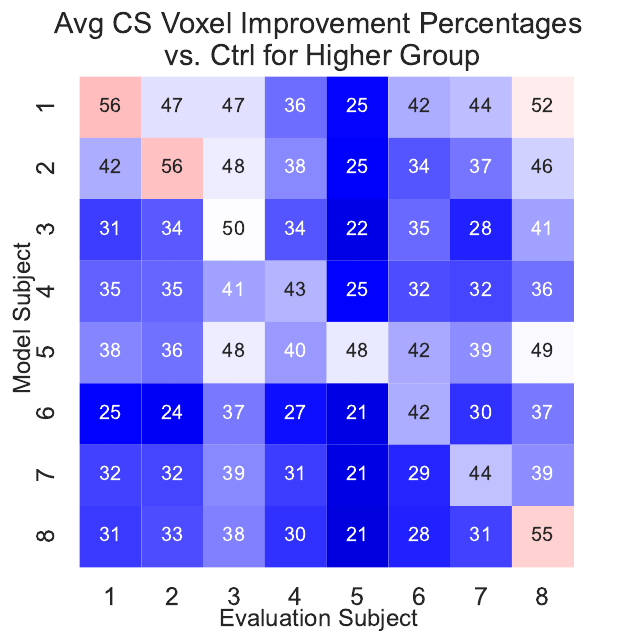}
    \caption{Cross-subject results for the early and higher visual ROI groups. The element in block $i,j$ is the percentage of improved when using subject $i$'s CL-tuned feature extraction model followed by PCA and $\ell_2$-penalized linear model to predict the fMRI responses for subject $j$ versus using features from the pretrained AlexNet CNN.}
    \label{fig:early rois cs avgs}
\end{figure}

To evaluate how well the CL fine-tuned feature extraction models transfer across subjects, we evaluate encoding performance using the cross-subject models described in Section~\ref{sec: cross subject models}. We visualize these results in a matrix, where the element in row $i$ and column $j$ is the percentage of voxels improved with CL fine-tuning versus the baseline pretrained AlexNet features across the test set samples for subject $j$ when using the model tuned using the fMRI responses from subject $i$. These matrices are created for each ROI that is present in all 8 subjects, and are then averaged across ROIs in one of the ROI groups (early or higher) to get the final cross-subject results for each ROI group. Figure \ref{fig:early rois cs avgs} shows the final cross-subject results by ROI group.

In general, we see subject-specificity among the CL-tuned models as the entries along the diagonal are often among the highest. For most subjects, on average across both ROI groups, the models trained using the image feature extraction tuned to the evaluation subject's data achieve the best performance relative to untuned AlexNet. However, this is not always true. 
For example, in subject $7$, we find that in the early visual ROIs, the models tuned on the data from subject $2$ achieve better performance ($80\%$ of voxels improved versus the untuned baseline when using the subject $2$ model versus $75\%$ of voxels improved versus the untuned baseline when using the subject $7$ model).  

ROIs in the early visual group have lower subject specificity than the other groups, with $5$ of the $8$ subjects seeing better performance with a model tuned on a different subject's data. In this group, the models trained on subject $1$ and $2$ consistently perform the best when transferred to the rest of the subjects. These subjects are the two subjects with the highest encoding accuracies for the early visual ROIs for the subject-specific models, and are also the subjects with the highest noise ceilings (see the NSD paper's supplementary information \citep{allen2022massive}). This suggests that the image processing across the same early visual ROI in different subjects is similar enough that subject-specific differences can be offset by using the responses from that ROI in different subjects where the data is less noisy. In the higher visual ROIs, where fine-tuning rarely improves, only subject 8 has more than half the voxels improved by subject 1's models.

\subsection{Pooled Subject Fine-tuning}
Pooling subjects while fine-tuning the AlexNet for a specific ROI creates a shared representation space that benefits from more data, and has the potential to improve the subsequent encoding for ROIs where cross-subject training succeeds. The encoding performance results with pooled models versus subject-specific models, summarized in Table~\ref{table:pooled_table_early} for the early group of ROIs and Table~\ref{table:pooled_table_higher} in \ref{sec: pooled results higher} for higher visual ROIs, echo the cross-subject transfer results: encoding performance is significantly improved when pooling (using the average embedding dimension) in early ROIs, but significantly worse when pooling in later ROIs.  For the early ROIs, the corrected p-value is $0.020$ and effect size is 1.34 (Cohen's d) for a paired, two-tailed t-test with Bonferroni correction for 3 tests and significance threshold of $\alpha=0.1$; and for higher ROIs, the corrected p-value is $0.00062$ with an effect size of 2.49. We also use the pooled models to test the effect of an average embedding dimension versus a constant embedding dimension (taken from the largest across all ROIs). In the early ROIs, the encoding model's average correlation is not significantly different with a corrected p-value of 0.24. Using a constant large embedding dimension yields a slight gain in the average proportion of improved voxels 69.1\% versus 67\% compared to the subject-specific approach. 
\begin{table}[H]
    \begin{tabular}{@{}cccccc@{}}
        \toprule
        \textbf{Sub} &
        \textbf{\begin{tabular}[c]{@{}c@{}}Sub-\\Specific \\ Avg. $\bar{\rho}$ \end{tabular}} &
        \textbf{\begin{tabular}[c]{@{}c@{}}Pooled\\ (h-avg) \\ Avg. $\bar{\rho}$ \end{tabular}} &
        \textbf{\begin{tabular}[c]{@{}c@{}}Pooled\\ (h-constant) \\ Avg. $\bar{\rho}$ \end{tabular}} &
        \textbf{\begin{tabular}[c]{@{}c@{}}Pooled (h-avg) \\ $>$ Sub-Specific \end{tabular}} &
        \textbf{\begin{tabular}[c]{@{}c@{}}Pooled (h-constant) \\ $>$ Sub-Specific \end{tabular}} \\ \midrule
        1 & {0.532} & {0.532} & {0.532} & 55.1$\%$ & 57.4$\%$ \\
        2 & 0.509 & 0.510 & 0.510 & 61.0$\%$  & 60.3$\%$ \\
        3 & 0.424 & 0.425 & 0.425 & 62.9$\%$  & 64.7$\%$ \\
        4 & 0.403 & 0.406 & 0.407 & 74.0$\%$  & 75.6$\%$ \\
        5 & 0.450 & 0.451 & 0.452 & 61.2$\%$  & 63.6$\%$ \\
        6 & 0.406 & 0.410 & 0.410 & {76.3$\%$}  & {81.4$\%$} \\
        7 & 0.406 & 0.410 & 0.410 & 70.6$\%$  & 70.5$\%$ \\
        8 & 0.342 & 0.346 & 0.347 & 75.0$\%$  & 79.3$\%$ \\ \midrule
        All & 0.434 & 0.436 & 0.437 & 67.0$\%$ & 69.1$\%$ \\ \bottomrule
    \end{tabular}
    \caption{Pooled versus subject-specific results for early visual ROIs. For each subject, the average test set correlation $\rho$ as described in Section~\ref{sec: accuracy metric} across all early visual ROIs is reported for CL-tuned AlexNet (Section~\ref{sec: contrastive learning model}) and pooled CL-tuned models using both a embedding layer dimension equal to the average voxel dimension of all subjects (h-avg) and an embedding layer dimension equal to the maximum voxel dimension over all subjects and ROIs (h-constant) (Section~\ref{sec: pooled models}). The last two columns show the percentage of voxels, averaged across early visual ROIs, for which encoding with the pooled model was more predictive compared to the subject-specific model.}
    \label{table:pooled_table_early}
\end{table}

\begin{figure}
    \centering
    \includegraphics{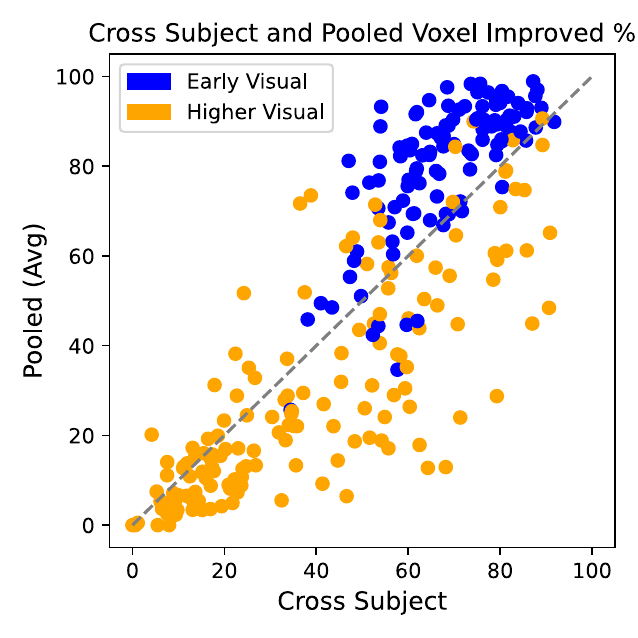}
    \caption{Comparison of percentage of voxels with  improved encoding when using CL fine-tuning versus the untuned AlexNet baseline for cross-subject models and pooled models (with averaged embedding layer dimension) across all subjects and ROIs. The average cross-subject improvement for a given ROI on the $i$th subject is calculated as the mean percentage of improved voxels versus the untuned baseline across all other subjects $j \neq i$ when using the $j$th subject's fine-tuned model. The pooled improvement is the percentage of voxels with higher correlation versus the untuned baseline for a given subject and ROI when using the pooled model for that ROI.}
    \label{fig:cs_vs_pooled_avg_scatterplot}
\end{figure}

In Figure \ref{fig:cs_vs_pooled_avg_scatterplot}, we compare the percentage of improved voxels versus untuned AlexNet for the pooled models and the cross-subject models across all ROIs, where for the cross-subject models for each ROI we average the percentage of improved voxels when using the models for all other subjects $j \neq i$ to predict subject $i$. We see a moderate correlation between the cross-subject and pooled model performance, which makes sense because the pooled models achieve higher performance in the same ROIs (early visual) where we see high cross-subject transferability. 

\subsection{External Dataset Validation}
Table~\ref{table:NOD-results} contains the results of the external dataset validation experiment on the Natural Object Dataset~\citep{gong2023large}, as described in Section~\ref{sec:NOD-method}. The  ROI-specific feature extraction models fined tuned using CL on NSD's Subject 1 improve the encoding prediction for voxels in the corresponding early ROIs in NOD.  For all 9 subjects, more than half of the top 50 voxels per ROI with the highest signal to noise ratio in the early visual areas (V1, V2, V3, and V4) are better encoded (higher correlation) with the CL-tuned. A subset of 4 of the 9 subjects have significant gains in average correlation ranging from 8.9\% to 15.9\% with $\ge$87.5\% voxels better encoded with the brain-tuned representation. Overall, the average proportion of voxels improved is $74.6\%$ and the average correlation goes from 0.127 to 0.136, an improvement of 7.1\%. This supports the conclusion that the CL fine-tuning creates feature extraction models that process visual stimuli more similar to the target ROI. 

\begin{table}[H]
    \begin{tabular}{@{}ccccccc@{}}
        \toprule
        \textbf{Subject} &
        \textbf{\begin{tabular}[c]{@{}c@{}}Avg. Ctrl. \\ $\bar{\rho}$ \end{tabular}} &
        \textbf{\begin{tabular}[c]{@{}c@{}}Avg. CL \\ $\bar{\rho}$ \end{tabular}} &
        \textbf{\begin{tabular}[c]{@{}c@{}}Avg. Reg. \\ $\bar{\rho}$ \end{tabular}} &
        \textbf{\begin{tabular}[c]{@{}c@{}}Avg. \% of \\voxels \\ improved \\vs. Ctrl.\end{tabular}} &
        \textbf{\begin{tabular}[c]{@{}c@{}}Avg. \% of \\voxels \\ improved \\vs. Reg\end{tabular}} &
        \textbf{\begin{tabular}[c]{@{}c@{}}CL vs. \\Ctrl. \\$\rho$ \\Change\end{tabular}} \\ \midrule
        1 & 0.153 & 0.157 & 0.144 & 56.5 & 78.5 & +2.6\%\\
        2 & 0.178 & 0.182 & 0.165 & 61.5 & 84.0 & +2.2\%\\
        3 & 0.124 & 0.137 & 0.132 & 88.5 & 70.5 & +10.5\%\\ 
        4 & 0.113 & 0.131 & 0.119 & 89.5 & 80.0 & +15.9\%\\ 
        5 & 0.121 & 0.128 & 0.124 & 72.0 & 66.5 & +5.8\%\\ 
        6 & 0.088 & 0.097 & 0.086 & 66.0 & 81.5 & +10.2\%\\ 
        7 & 0.117 & 0.131 & 0.126 & 94.0 & 74.5 & +12.0\%\\ 
        8 & 0.146 & 0.159 & 0.156 & 90.5 & 65.0 & +8.9\%\\ 
        9 & 0.100 & 0.097 & 0.098 & 52.5 & 51.0 & -3.0\%\\ \midrule
        All & 0.127 & 0.136 & 0.128 & 74.6 & 72.4 & +7.1\%\\ \bottomrule
    \end{tabular}
    \caption{Encoding performance for early visual ROIs (V1-V4) for subjects 1-9 in the Natural Object Dataset using feature extraction models tuned on NSD as described in Section~\ref{sec:NOD-method}. For each subject, the average test set correlation $\rho$ as described in Section~\ref{sec: accuracy metric} across all early visual ROIs is reported for untuned AlexNet, the AlexNet models tuned using CL on NSD, and the AlexNet models tuned using regression on NSD. The percentage of voxels for which the CL-tuned model was more predictive on average across the test set compared to the untuned AlexNet and regression-tuned AlexNet is also shown.}
    \label{table:NOD-results}
\end{table}

The average encoding correlations for the NOD dataset are much lower than for the early visual ROIs in the NSD datasets (the best subject's baseline is 0.178 in NOD versus 0.524 in NSD). This is not unexpected due to three reasons: first, the scanner resolution for NOD is lower than NSD (3T instead of 7T); second, the responses to each image for NOD are only the result of a single trial rather than an average of 3 trials as in NSD; and third, NOD has fewer responses used for fitting the linear models than NSD (3400 instead of approximately 9000 per subject). However, the relative encoding performance for different methods still provides a meaningful metric for how well each feature extraction model is tuned for that ROI. Because the CL- and regression-tuned models were fine-tuned on a completely different dataset, the improved encoding accuracy for the brain-tuned models over the untuned models indicates that fine-tuning on NSD generalizes to other datasets. 

\subsection{Downstream classification tasks}
\begin{figure}[!htb]
    \centering
    \includegraphics[width=.45\textwidth]{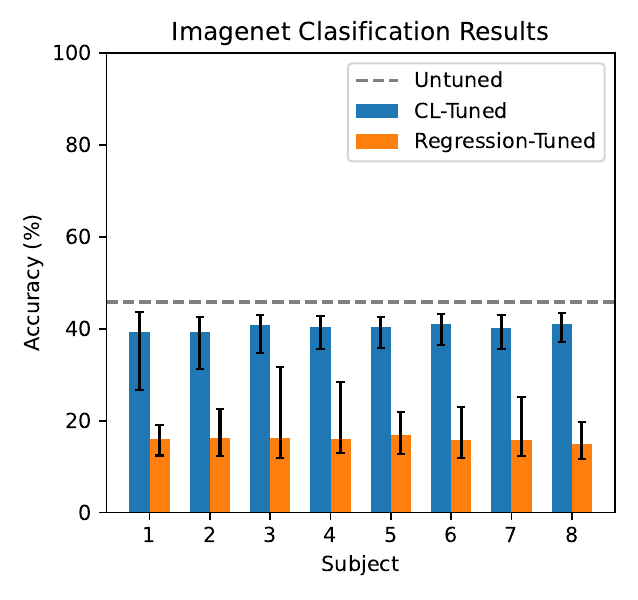}
    \\[\smallskipamount]
    \includegraphics[width=.45\textwidth]{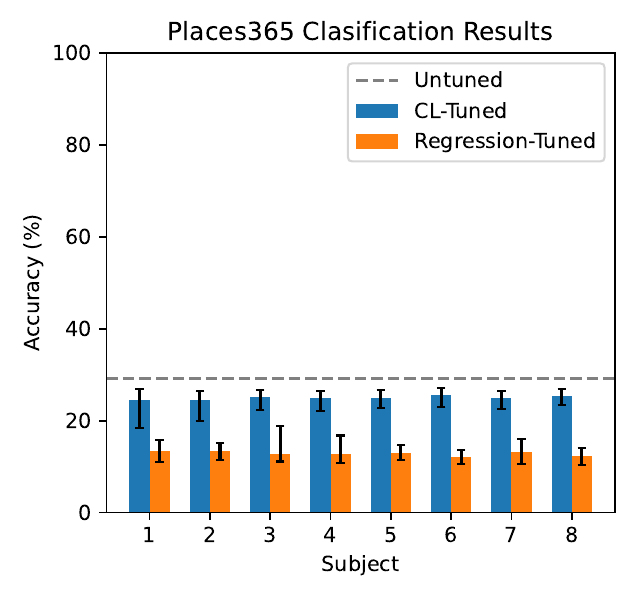}
    \includegraphics[width=.45\textwidth]{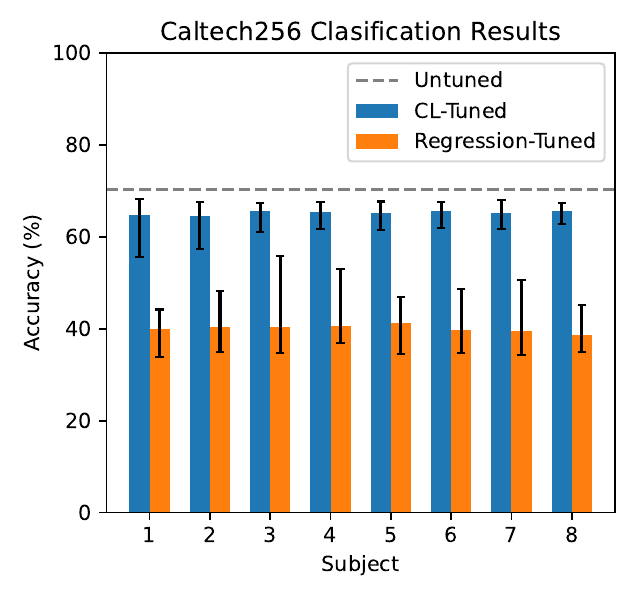}
    \caption{Results for baseline and brain-tuned AlexNet on the validation portion of ImageNet  ILSVRC2012 (the training portion was used for the pretraining the AlexNet baseline), Caltech256, and Places365. The bar height is the average accuracy across all ROIs tuned using that method for that subject. Error bars represent the full range of results across all ROIs (minimum to maximum).}
    \label{fig: image classification results}
\end{figure}

Using the methodology from Section~\ref{sec: downstream tasks} we evaluate the test set classification accuracy for image classifiers built on top of the untuned, subject-specific CL-tuned and regression-tuned models for the ImageNet ILSVRC2012, Caltech256, and Places365 datasets.  Figure \ref{fig: image classification results} summarizes the performance of the subject specific tuned models versus the untrained baseline. See Figures \ref{fig:caltech256_results_mat}, \ref{fig:places365_results_mat}, and \ref{fig:imagenet_results_mat} in ~\ref{sec: downstream appendix} for the classification accuracy per ROI for both the subject-specific and pooled models for each ROI.  

For both the ImageNet classification task and the non-training classification tasks (Caltech256 and Places365), we see a drop in classification accuracy when using the brain-tuned models, but the CL-tuned models suffer much less of a decrease than the regression-tuned models. For ImageNet, the baseline pretrained on ImageNet AlexNet achieves 45.8\% accuracy. Across all possible subject-ROI pairs, the best CL model achieves a test accuracy of 43.6\% (LH EBA in subject 1), see Figure~\ref{fig:imagenet_results_mat}. This means there is only a 1.6\% drop in performance after tuning to brain-activity (while starting from pretrained).
For Caltech256, the baseline pretrained model achieves 70.3\% accuracy. Across all possible subject-ROI pairs, the best CL model achieves a test accuracy of 68.2\% (LH EBA in subject 1), see Figure~\ref{fig:caltech256_results_mat}.
For Places365, the most challenging task, the baseline pretrained model achieves 29.1\% accuracy.  Across all possible subject-ROI pairs, the best CL model achieves a test accuracy of 27.2\% (LH OPA and LH PPA in subject 6), see Figure~\ref{fig:places365_results_mat}. 
These results show that on average, the CL-tuning preserves most of the information content necessary for using the image features of the CNN for image classification tasks, especially in task-relevant ROIs. Quantitatively, the values are consistent across subjects. 

\subsection{Model Landscapes}
To understand the similarity of the classification models trained for the ImageNet dataset on top of the feature extraction models we apply multidimensional scaling (MDS) of the Bhattacharya based dissimilarity metric~\citep{mao2024training} between the prediction vectors from the downstream classifier for ImageNet for each model. When we apply this to the full set of models (untuned AlexNet, CL-tuned for each subject and ROI, CL-tuned pooled subject for each ROI, and regression tuned for each subject and ROI), we find that each group of models is distinct, as shown in Figure~\ref{fig:overall_landscape} in ~\ref{sec: model landscapes appendix}. In comparison, models landscapes with different subsets of the models will use different components of the dissimilarity space as the axes of model embedding coordinates.

\begin{figure}
    \centering
    \includegraphics[width=0.53\linewidth]{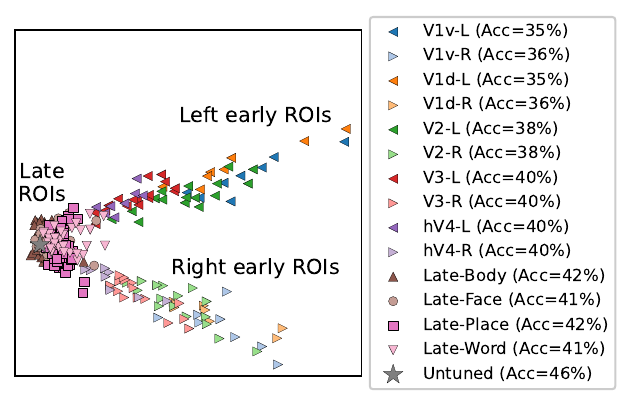}
    \includegraphics[width=0.425\linewidth]{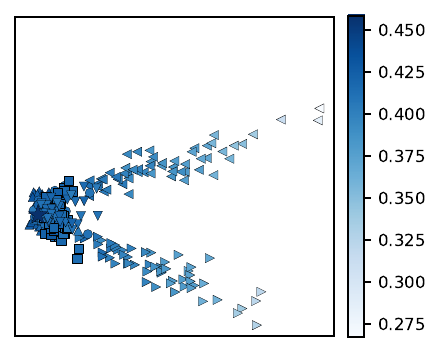}    
    \caption{Model  landscape for untuned AlexNet (star) and  subject-specific ROI-specific CL fine-tuned models. The early visual ROIs are separated into left and right, with early ROIs having lower classification accuracy (right).}
    \label{fig:subject_landscape_key}
\end{figure}

Figure~\ref{fig:subject_landscape_key} shows the landscape when taking only the untuned AlexNet and subject- and ROI-specific CL fine-tuned models. Models tuned with CL for different subjects, but the same ROI, appear in the same arrangement and are relatively similar based on the their embedding coordinates. The landscape is arranged such that the first axis separates early versus late ROIs, but the second axis separates the left and right hemispheres in the early ROIs (V1–4). This indicates that predictions for left and right ROI models are systematically different across the images—possibly due to the lateralized presence of objects related to the classification task and the retinotopic tuning in V1–4, which we investigate in Figure~\ref{fig:gradcam}. Higher level (late) ROIs have more similar predictions to untuned AlexNet predictions, and are more accurate.  

\begin{figure}[H]
    \centering
    \includegraphics[width=0.5\linewidth]{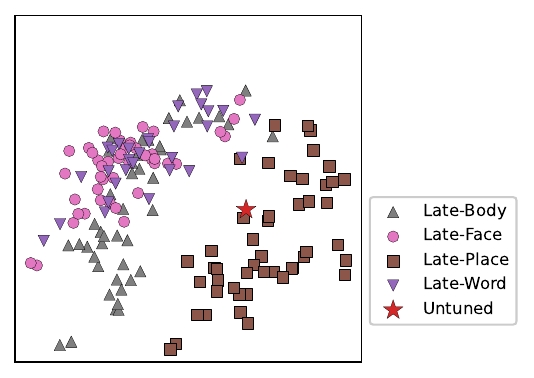}
    \includegraphics[width=0.45\linewidth]{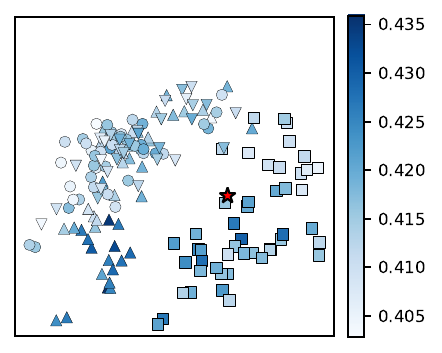}
    \caption{Model landscape for untuned AlexNet (star marker) and higher level (late) subject-specific ROI-specific fine-tuned models. }
    \label{fig:higher}
\end{figure}

Figure~\ref{fig:higher} shows the landscape when taking only the untuned AlexNet and subject- and ROI-specific CL fine-tuned models for higher level ROIs. Here we see the two axes separate the models by ROI group related to representation of body, face, place, and words.  The ROIs associated to face and place are generally worse for classification, but clearly distinct in this embedding. Again, models tuned for different subjects but the same ROI appear in the same arrangement and are relatively similar based on the their embedding coordinates.  Notably the untuned AlexNet is not most similar to the highest performing fine-tuned model: subject 1's EBA (left hemisphere), as detailed  in Figure~\ref{fig:imagenet_results_mat}. This reveals that fine-tuned models may have deviations in the predictions from the untuned AlexNet while still retaining features useful for downstream classification tasks. 

Figure~\ref{fig:pooled_landscape_key}, in \ref{sec: downstream appendix},  shows the landscape when taking only the untuned AlexNet and pooled-subject, ROI-specific CL fine-tuned models matches the landscape for the subject-specific CL fine-tuning. 

Finally, we note that the landscape for the regression-tuned, subject-specific and ROI-specific models differs from the CL-tuned model landscapes, as shown in Figure~\ref{fig:reg_landscape} in \ref{sec: downstream appendix}. Notably, while the ROIs of different subjects are similar and the primary axis of dissimilarity is associated with early versus later ROIs, the second axis separates place and face ROIs in the later areas. Looking at only the later ROIs, there is less clustering than in the CL tuning. This indicates that although regression tuning does create ROI-specific feature extraction, it does not yield as consistent representations as CL across the subjects, and does not reveal any differences between the information in left and right hemispheres for early visual ROIs.

\subsection{Salience Maps for CL-tuned Models}
\begin{figure}
    \centering
    \includegraphics[width=0.99\linewidth]{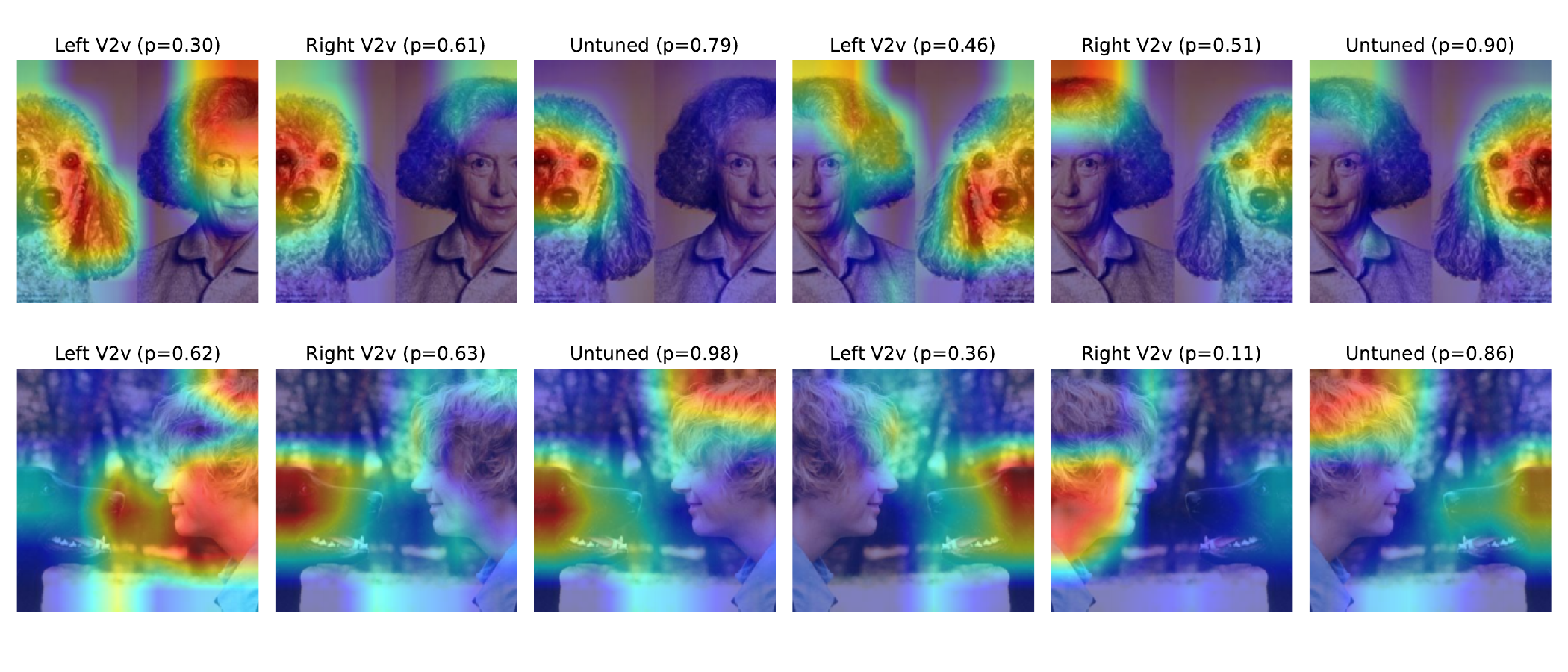}
    \caption{Grad-CAM salience maps and predicted probability for the 'dog' class ($p$) on two Caltech256 images, with and without horizontal flipping, for left and right V2v-tuned models and untuned AlexNet.}
    \label{fig:gradcam}
\end{figure}

To illustrate how the CL-tuned models can reveal visual processing of different visual ROIs, we investigate the presence of lateralization in the tuning of early visual ROIs through images with asymmetric content. As described in Section~\ref{sec:saliency methods}, Grad-CAM salience maps~\citep{selvaraju2017grad} are generated for the untuned AlexNet and for CL-tuned models (subject 2's left-V2v and right-V2v) with classification heads trained on the Caltech-256 dataset. Figure~\ref{fig:gradcam} shows the Grad-CAM salience maps and predicted probability for two images with the label `dog', which have a person on the opposite side, with and without horizontal flipping.  We observe that for the right hemisphere V2v-tuned model, the predicted probability for the `dog' class increases when the dog is on the left side of the image for both images, and has higher salience on the left side even when it is not the class of interest. For the left hemisphere V2v-tuned model, the predicted probability increases when the dog is on the right side of the image for the top image, and has higher salience for the right side even when it is flipped; the salience map for the bottom image is more focused on the dog only when it on the right side. In comparison, the salience maps of the untuned AlexNet with a retrained classification head are more symmetric. This demonstration points to the potential for future work to use a more systematic experiment to investigate consistency of lateralization effects for early visual ROIs, as well as effects for other ROIs, such as whether a model tuned using CL for a face-selective ROI is more predictive of faces, or whether a brain-tuned region highlights different visual features compared to standard pretrained networks. 

\section{Discussion}
We have shown that the image feature representations extracted from convolutional neural networks pretrained on image classification tasks can be made more predictive of fMRI activity in the human visual cortex in the early visual regions by fine-tuning using a contrastive learning (CL) approach~\citep{gutmann2010noise}  that maximizes a lower bound on the mutual information~\citep{poole2019} between the fMRI activity and the image features. This required a novel adaptation of the SimCLR approach~\citep{chen2020simple} to pair disparate spaces of natural and artificial neural responses, similar to the CLIP model~\citep{radford2021learning} trained to pair images with captions.  Essentially, the CNN's processing is modified such that the representations of images is better aligned to human subject's brain's representation of the same images. We believe this approach is a natural evolution of neural encoding models that seek to capture the dependency of stimulus and neural response \citep{meyer2017models}. 

Our approach of fine-tuning ROI- and subject-specific feature extraction models using CL: 1) improves a majority of the voxels for each subject of the voxel’s linear encoding on the Natural Scenes Dataset~\citep{allen2022massive}; 2) achieves much better results than comparable regression-based tuning; 3) exhibits cross-subject transferability for early visual ROIs with improvements in encoding on the majority of 9 subjects from another lower-resolution dataset~\citep{gong2023large}; and 4) reveals divergence from the original CNN in terms of performance and similarity of prediction on downstream tasks across subjects and ROIs with meaningful interpretation of ROI-specific model similarities. 

While the encoding of ROIs earlier in the visual processing stream are consistently improved when using the fine-tuned models, ROIs in higher visual ROIs such as face- and body-selective regions have minimal improvement over the untuned AlexNet baseline (average proportion of voxels in higher visual ROIs  improved on only 3 of the 8 subjects). We hypothesize that this may be because the higher visual ROIs have more specialized roles in visual processing, such as being selective for faces or words, which may not be well-represented in the Natural Scenes dataset used here \citep{szwed2011specialization, kanwisher1997fusiform}.  If an ROI has a highly-specialized purpose, then there may be a limited number of images in the training set that are encoded in a meaningfully different way by the ROI and thus not as many useful training images as in the early visual ROIs, which encode lower level image features that should vary more across images. Alternatively, the untuned AlexNet may already provide useful features, and CL-tuning does not improve it further. 

In early visual ROIs our CL approach significantly improves the prediction (in terms of average correlation) versus baseline of pretrained AlexNet for the NSD dataset with a sample size of $n=8$ (paired, one-tailed t-test with Bonferroni correction for 2 tests and significance threshold of $\alpha=0.1$). While the effect size is $1.65$, it is a relatively small average improvement in the average correlation, $0.430$ to $0.434$ (only $0.9$\%). The highest average correlation is seen in Subject 1 with an improvement of 1.5\%. Nonetheless, gains exist across a majority of the voxels in the early visual areas. An average of $75.5\%$ of voxels are improved when compared to the predictions of the baseline pretrained AlexNet (which therefore means worsening the prediction in $24.5\%$ of voxels).  In comparison, an alternative approach that modifies the CNN's weights by directly minimizing a regression loss for the encoding actually lowers the average correlation of the prediction to $0.424$ due to overfitting. On average, $89.5$\% of the voxels are better encoded using the CL fine-tuning compared to regression fine tuning.  This suggests that CL has utility in helping learn representations that are more robust in terms of their similarity to target neural responses.

Importantly, feature extraction models tuned to the fMRI responses to a specific subject in an early visual ROI can be used in the same ROIs in other subjects with minimal drop in encoding performance,  suggesting that the learned feature representations in the fine-tuned AlexNet are common across the different subjects for a particular ROI. (This finding is also confirmed by similarity-based embedding in the model landscapes.) Interestingly, the early group of ROI tuned models from subjects 1 and 2 are more predictive of the responses for all other subjects than those subject's model tuned specifically on their data. In contrast, for the higher visual groups, nearly all of the subjects' test responses were best predicted by the models tuned on their own training data. This suggests that the visual representations in early visual ROIs (V1-V4) across different subjects may be more similar than the representations in later visual group ROIs. Previous work by \citet{wang2015probabilistic} showed that higher inter-subject variability exists in the structure of higher visual ROIs than ROIs in early visual cortex, which could explain why the models tuned on early visual ROIs are more transferable to different subjects. To further test this, we trained pooled models that combine data from multiple subjects to create a more general model that could benefit from the scaling of training data that is possible when using data from multiple subjects. We found that pooling enables feature extraction with significantly better encoding in the early group of ROIs, but significantly worse in later group. 

We further validated our approach by applying the fine-tuned feature extraction models tuned on the NSD dataset to 9 subjects from the Natural Object Dataset~\citep{gong2023large} for early visual ROIs. Notably, the dataset is lower resolution (3T versus 7T) and only 50 voxels per ROI with the highest noise ceilings are used.  We fit new linear models using these models to predict the fMRI responses in NOD, using the same preselected layer per ROI, but find a new penalty term $\alpha_r$. We find that an average of $74.6\%$ of voxels are improved versus the untuned AlexNet baseline across all early visual ROIs and subjects when using the CL-tuned models fit on the NSD dataset. With the CL-tuned models the average correlation goes from 0.127 to 0.136, an improvement of 7\%. A subset of 4 of the 9 subjects have meaningful gains in average correlation ranging from 8.9\% to 15.9\%. This finding further suggests that the CL fine-tuning on specific ROIs can help in lower resolution datasets. Finally, it confirms that the CL fine-tuned AlexNet backbone extracts visual features more predictive of the early visual ROIs.

Beyond encoding, we use downstream classification tasks to examine the performance of the set of CL-tuned AlexNet models. While all of the brain-tuned models experience small to moderate decreases in accuracy compared to the untuned (but pretrained on ImageNet) AlexNet, models tuned to higher level ROIs (EBA, OPA, PPA) have minimal drops in accuracy on downstream classification tasks. Notably, the CL-tuned models have much better performance than the regression tuned models, which also performed worse for encoding. Interestingly, the pooled models had worse downstream performance but exhibit a landscape and relative pattern of performance across ROIs that matched the subject-specific models. This shows that the CL-tuning with more brain data may further deviate from the pretrained AlexNet yielding better encoding performance (Table~\ref{table:pooled_table_early}) but worse downstream task performance. Holistically, across ROIs the CL-tuned models with highest encoding accuracy (pooled early visual) had the lowest downstream classification. This makes sense as the classification tasks require higher-level features. In terms of the similarity-based embedding in the model landscapes, pooled models initially appear distinct from subject-specific models due to the increased dataset size creating further fine-tuning (see Figure~\ref{fig:overall_landscape}); however, examining the pooled and subject-specific models separately reveals that the relative similarity among the ROIs are matched (see Figures~\ref{fig:subject_landscape_key}, \ref{fig:higher}, and \ref{fig:pooled_landscape_key}).  The two key features of this landscape are  that the the CL-tuned models are organized primarily by early to late, with the late models being more similar to untuned, and that the ROIs for the left and right hemispheres of early visual ROIs are segregated. A possible explanation is that since the subjects in the NSD dataset fixated on the center of images, there are lateralized effects present in the fMRI response due to the retinotipic mapping of early areas combined with the presence of features or objects appearing on one side of the image or the other. The CL-tuning may have captured this effect creating asymmetric image feature extraction. To investigate this, we used salience map based explainability methods, namely  GradCAM  \citep{selvaraju2017grad}, and found preliminary evidence of lateralization in the saliency maps in left and right ventral V2. A similar approach could be leveraged to investigate other substantive differences between  brain-tuned feature extraction models and pretrained networks, enabling further ``in-silico'' experiments.

\subsection{Limitations}
We note that a number of benchmarks were set for encoding on the Natural Scenes Dataset (NSD)~\citep{allen2022massive} based on the Algonauts Project 2023 competition~\citep{gifford2023algonauts}. We do not claim to achieve state of the art, but rather we use NSD as a uniquely large dataset to understand how contrastive learning can yield ROI and/or subject-specific tuning of feature extraction models. Our contribution is a method to compare ROI-specific fine-tuned models by examining their transferability and their similarity and performance on different downstream tasks. Likewise, we have used the Natural Object Dataset~\citep{gong2023large}, which uses a different parcellation scheme~\citep{glasser2016multi} than the manual ROI definitions in NSD, to confirm that the transferability of the feature extraction models tuned to early visual ROIs which outperform the baseline, but do not directly compare to the previous encoding experiments~\citep{gong2023large} that using population receptive field models for retinotopic mapping and $\ell_1$ regularized linear regression. 

The design choices we have used may not be optimal. In particular, fine-tuning the entire AlexNet and then using a pre-selected layer (based on untuned AlexNet acting as a feature extraction for encoding) does not directly ensure that the activation layer used for linear encoding is optimal; however, this approach of discarding the subsequent layers acts like the so-called ``guillotine regularization''~\citep{bordes2023guillotine}. We have also selected a ROI-wide choice of ridge regression penalty $\alpha_r$ based on a single subject for simplicity, but subject-specific voxel-wise selection could yield further improvements. Similarly, using ridge regression is not ideal as it does not account for the correlation of neighboring voxels. Partial least squares or rank constraints across the weight matrices can further improve the encoding~\citep{ranjbar2024structurally,krishnan2011partial}.

The choice of a nonlinear projection head, while used in SimCLR~\citep{chen2020simple}, is not used by CLIP~\citep{radford2021learning}. \citet{chen2020simple} and \citet{bordes2023guillotine} both showed that more information may be preserved if a non-linear projection head is used, but then dropped. However, for our purposes the choice of a dimension reducing linear projection and/or non-linear projection head on the voxel response is questionable since all of the information in the response is relevant. It may be the case that with the linear projection (of reduced dimension) and non-linear projection head the contrastive learning optimizes the model for a subset of the voxels in an ROI that best aligns with the feature extraction. This could explain why the encoding results are not improved across all voxels. For linear encoding, using the neural response and network activations directly (or through a variance preserving projection) may be more optimal. Nonetheless, it is necessary to have at least one linear map so that the fMRI response and AlexNet embedding are of the same dimension. Future work should examine the benefit of dimensionality reduction and/or the non-linear projection head.  

Additionally, our choice of scaling the embedding dimension in the contrastive learning loss to be proportional to the voxels in an ROI for the subject-specific ROI-specific models may not be optimal.
With the pooled tuning, we tested whether  using a constant embedding dimension in the CL loss performs better than an embedding dimension proportional to the number of voxels. While not significant in terms of encoding performance, the constant embedding dimension had marginally better performance in terms of both average test set correlation and percentage of voxels improved. This means that the particular choice of proportional scaling may not be an optimal hyper-parameter, but is not degrading performance and does not affect any of the conclusions. 

Similarly, we have used a fixed temperature for the loss function. Although the temperature is in the range of prior work~\citep{chen2020simple} it may not be optimal and  may limit the lower bound estimate of mutual information. In contrast, CLIP~\citep{radford2021learning} treats the temperature as a learnable parameter. 

While we tuned a separate model for each ROI, it could be argued that one could use a whole brain approach that tunes a single feature extraction model useful for predicting all ROIs. However, it is not clear how the subsequent analysis of this feature encoding would be approached. Instead, we explicitly disentangle the representation by creating a differently fine-tuned model for each ROIs. The benefit of our approach is that we can analyze the ROI-specific models. However, a limitation of our approach is the computation required—which has limited our hyper-parameter searches.

Future work could, firstly, fine-tune the image feature extraction using information pooled across the fMRI responses across different regions of the visual processing stream (i.e. a whole-brain contrastive learning), and secondly, incorporate cross-attention mechanisms to use different ROIs to tune different parts of the CNN, similar to the approach of \citet{st2023brain} used to build models predictive of V1-V4, but extended to more ROIs. More generally, the contrastive learning approach could be used with neural data from other experimental modalities to tune other deep neural network architectures, such as fMRI responses to spoken or read text being used to modify language models.

\section{Conclusion}
In this work, we developed a novel methodology for neural encoding, based on using a contrastive loss to fine-tune a neural network to maximize a lower bound on the shared information between the fMRI responses in visual regions of interest (ROIs) and the stimulus images of natural scenes. We find a significant improvement in encoding performance across early visual ROIs, but not in higher visual cortex areas identified as being selective for words, faces, bodies, or places. The representations learned by fine-tuning for individual subjects tend to be transferable as feature extractors in the early visual ROIs, including to encoding the fMRI responses to natural images in subjects from other datasets. We use classification performance of classifiers built on top of fine-tuned models on both ImageNet, the original training task for the untuned AlexNet, and two other image classification datasets (consisting of objects and places), to understand how much the fine-tuned networks deviate from the untuned network. We use the similarity of predictions to create model landscapes to visualize the similarity of responses, revealing multiple axes of dissimilarity associated to early versus late ROIs, lateralization in early ROIs, and separation of place from body and scene areas. We highlight preliminary evidence using salience maps to support the lateralization in the brain-tuned models for early visual ROis. We found that pooling together neural responses from different subjects to scale up the training dataset used for fine tuning further improves encoding performance in early visual areas, and creates more deviations from the untuned model, but reveals the same axes of dissimilarity as the subject-specific tuning. Overall, this works points to the potential of aligning   different layers of one neural network to different brain regions, and the use of explainability methods to examine differences in image processing between the untuned and brain-tuned versions of the neural network.

\section*{Acknowledgements}
This research was carried out with the support of the University of Delaware Research Foundation. This research was supported in part through the use of Information Technologies (IT) resources at the University of Delaware, specifically the high-performance computing resources. The authors acknowledge the thoughtful reviewers who helped in improving the manuscript. 

\begin{appendix}
\renewcommand\thesection{Appendix~\Alph{section}}
\section{ROI Definitions}
\label{appA}
\begin{figure}[H]
    \centering
    \includegraphics[scale=1.25]{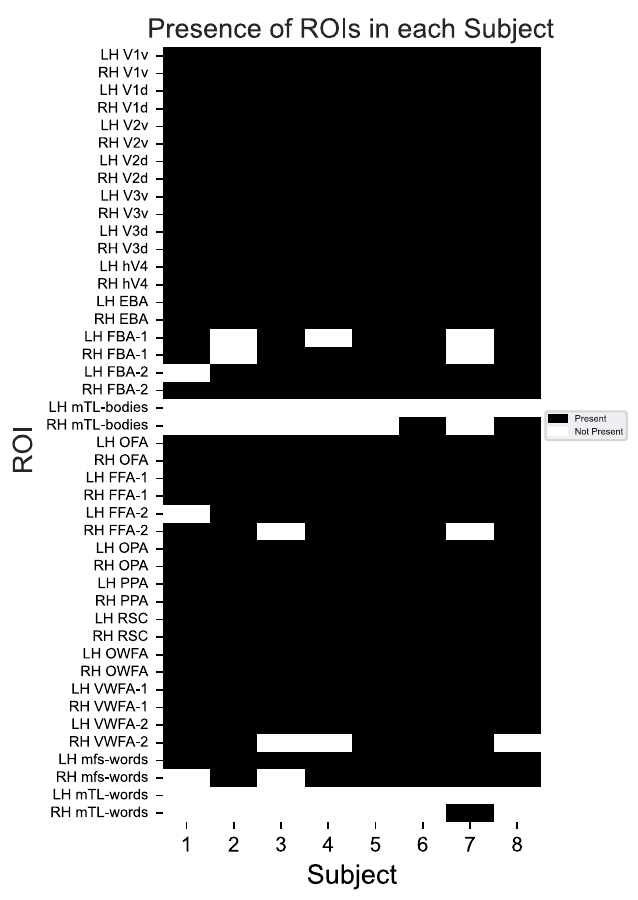}
    \caption{ROIs present in each subject from NSD, with black indicating that the ROI is present in the subject, and white indicating that the ROI is not present.}
    \label{fig:roi_subj_presence_table}
\end{figure}

ROIs are characterized according to 5 main groups: early visual regions, body-selective regions, face-selective regions, place-selective regions, and word-selective regions. The early visual regions (V1v, V1d, V2v, V2d, V3v, V3d, hV4) were manually defined based on population receptive field mapping (pRF) experimental results. The body- (EBA, FBA-1, FBA-2, mTL-bodies), face- (OFA, FFA-1, FFA-2, mTL-faces, aTL-faces), place- (OPA, PPA, RSC), and word-selective regions (OWFA, VWFA-1, VWFA-2, mfs-words, mTL-words) were manually created using the results of a functional localizer (fLoc) experiment. For more detailed information, see the Natural Scenes Dataset paper~\cite{allen2022massive}, especially the supplementary information. \\
Not all ROIs are present in all of the subjects. For the purpose of computing averages across subjects for a particular ROI, those subjects who did not have that ROI were ignored. Table \ref{fig:roi_subj_presence_table} illustrates what ROIs are present in each subject.

\section{Estimation of Mutual Information Lower Bound}
\label{sec:mi}
For continuous random variables $X\in\mathcal{X}$ and $Y\in\mathcal{Y}$ which are jointly distributed, with independent and identically distributed copies of size $K$, $\{(X_i,Y_i)\}_{i=1}^K$, the mutual information between the variables $I(X,Y)$ is lower bounded by the following noise-contrastive estimation (NCE) where $\omega: \mathcal{X}\times\mathcal{Y}\rightarrow \mathbb{R}$ is a {continuous} function~\citep{poole2019}, 
\begin{align}
\label{simclr_mi}
    I(X,Y) \ge \sup_\omega \mathbb{E}[ \frac{1}{K} \sum_{i=1}^K   \log \frac{e^ {\omega(X_i,Y_i) }}{\frac{1}{K} \sum_{j=1}^K e^{ \omega(X_i, Y_j) }  }]
\end{align}
NCE's lower bound is itself upper-bounded by $\log (K)$, where $K$ is the batch size. Thus, cases of relatively large values of mutual information require a large batch size.  {Any mutual information estimate is limited and \eqref{simclr_mi} is conservative in practice even when $I(X,Y)\le \log(K)$~\citep{mcallester2020formal}.}  

{In the case of neural responses to natural images,} the fine-tuned feature extraction model $f_A^r$, the linear mappings to a shared latent space $\mathbf{W}^{x,r},\mathbf{W}^{y,r}$, and the non-linear projection head $g^r$ yield the function  $\omega:\mathcal{I}\times\mathcal{Y}^r\rightarrow [0,\frac{1}{\tau}]$
\begin{equation}
    \omega( x ,\mathbf{y}^r)=\frac{1}{\tau} \mathrm{sim}(\mathbf{z}^{x,r},\mathbf{z}^{y,r} )=\frac{1}{\tau} \mathrm{sim}(g^r( \mathbf{W}^{x,r} f_A^r(x) ), g^r( \mathbf{W}^{y,r} \mathbf{y}^r) ),
\end{equation} 
which can also be interpreted as a non-linear similarity function between the fine-tuned AlexNet activations $\mathbf{a}^r=f_A^r(x)$ and the ROI's voxel activations $\mathbf{y}^r$.  Theoretically, the optimal function is $\omega^*(x,\mathbf{y}^r) = \log p(\mathbf{y}^r|x) + c(\mathbf{y}^r)$ where $p(\mathbf{y}^r|x)$ is the conditional density and $c(\mathbf{y}^r)$ is any function that does not depend on $x$~\citep{ma2018noise}. If $\omega$ is close to optimal and $\mathbf{y}^r$ is the response to image $x$, then the vectors $\mathbf{z}^{x,r}$ and $\mathbf{z}^{y,r}$ should have a relatively high cosine similarity (small angle) yielding a relatively high conditional density $\log p(\mathbf{y}^r|x)$. If $x$ and $\mathbf{y}^r$ are independent then the vectors should have low cosine similarity on average with low conditional density $\log p(\mathbf{y}^r|x)$.

\begin{figure}[H]
    \centering
    \includegraphics[scale=1.15]{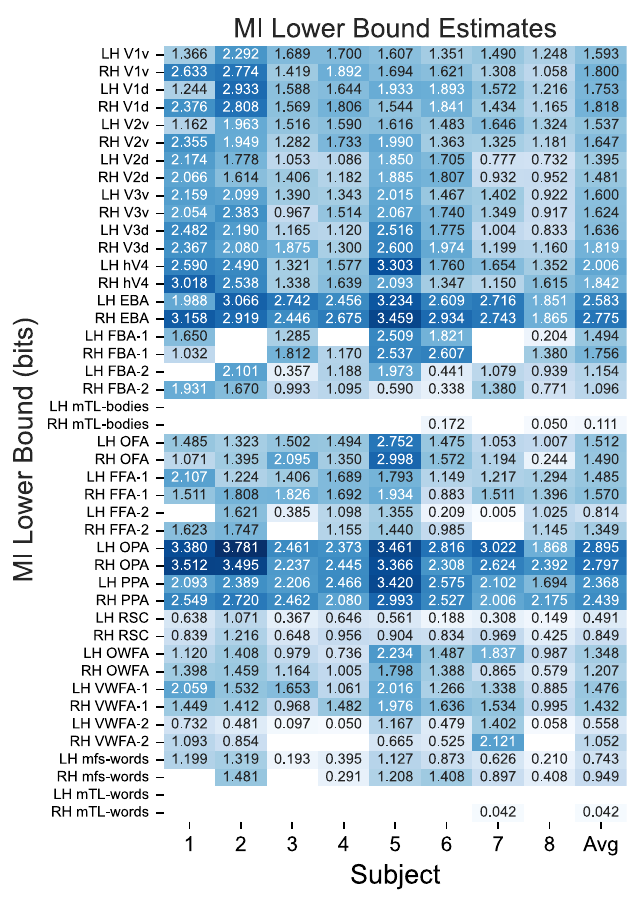}
    \caption{Mutual information lower bound estimates for each subject and ROI, in bits. The rightmost column is an average of the estimates for all subjects which have that ROI. Blank entries indicate that the ROI given by the row is not present in the subject given by the column.}
    \label{fig:mi_matrix}
\end{figure}

After training, the CL loss function enables the computation of a lower bound on mutual information between the image input as represented by the fine-tuned AlexNet and the fMRI activity. While potentially loose~\citep{mcallester2020formal}, we compute estimates of the lower bounds on the mutual information $I(X, Y)$ between the CL-tuned AlexNet representations and the fMRI responses using \eqref{simclr_mi} for all ROIs across all subjects. Figure~\ref{fig:mi_matrix} shows the lower bound estimate for CL-tuned subject-specific and ROI-specific model and CL-tuned ROI-specific models, as well as an average for each ROI over all subjects that have that ROI. 

There is a large gap before the theoretical upper bound of this lower bound determined by the batch size ($\log_2(1024)=10$ bits), and the largest estimate ROI mutual information of 3.78 bits for OPA (left hemisphere) in subject 2.  Nonetheless, the relative mutual information across the subjects and ROI specific models shows generally higher information in higher level areas (EBA, OPA, PPA). Some subject showed relatively high mutual information in lower level areas such as V1 and hV4.

\begin{figure}[H]
    \centering
    \includegraphics{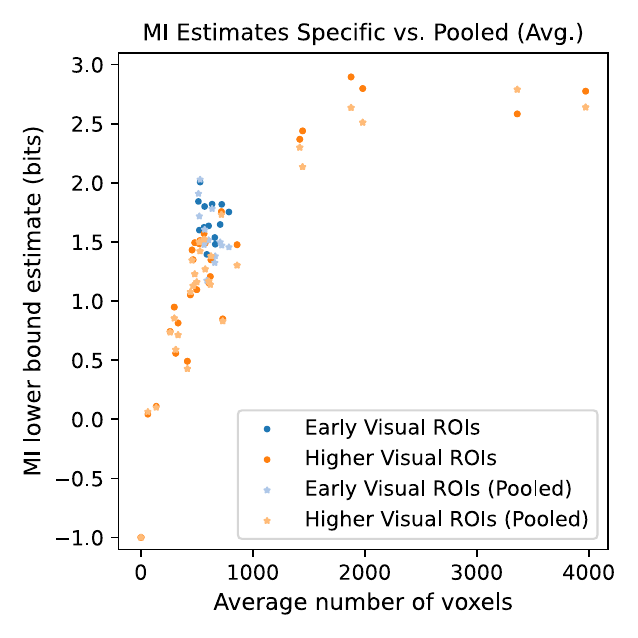}
    \caption{Lower-bound estimates on MI between the fMRI responses in each ROI and the corresponding ROI-specific CL-fine-tuned AlexNet output versus the average number of voxels in that ROI across subjects. For the subject-specific models the average MI is plotted. For the pooled models the embedding layer dimension is based on the average of the voxel dimensions of all subjects with the ROI. ROI groups are color coded.}
    \label{fig:mi_havg_plot}
\end{figure}

Figure~\ref{fig:mi_havg_plot} shows a plot of the average lower bound estimate across subjects for each ROI versus the average number of voxels in that ROI across the subjects, for both the subject-specific models and the pooled models when using the average embedding layer dimension. Generally, larger ROIs can carry more information, but this saturates above 200 voxels. The pooled models appear to have roughly the same information compared to the subject-specific models. 

While increasing mutual information with dimensionality makes sense, it is not clear if the relatively low values of the MI estimate and the saturation effects are due to limitations of the NCE bound, or whether the noise in fMRI causes the information content to be limited.  To investigate this we used the classification accuracy on ImageNet's 1000 classes and Fano's inequality~\citep{scarlett2021introductory} to lower bound the mutual information. Using the accuracy on ImageNet and  Fano's inequality the information content related to the class labels of the brain-tuned image feature representation must be higher than the mutual information between the representation and the neural response. For example, with a uniform distribution across ImageNet's 1000 classes Fano's inequality based on the accuracy~\citep{scarlett2021introductory} yields a lower bound of 3.3 bits for subject 1 EBA (left hemisphere), compared to 2 bits from the NCE estimate shown in Figure~\ref{fig:mi_matrix}. Thus, future work is needed to explore better mutual information estimation techniques~\citep{mcallester2020formal} to fMRI responses to stimuli.

\section{Pooled Results for Higher Visual ROIs} \label{sec: pooled results higher}
In Table~\ref{table:pooled_table_higher}, we show the test set results when using the pooled models as compared to the subject-specific models in the higher visual ROIs (where we see no benefit in subject-specific encoding accuracy on average with CL-tuning).
\begin{table}[H]
    \begin{tabular}{@{}cccccc@{}}
        \toprule
        \textbf{Sub} &
        \textbf{\begin{tabular}[c]{@{}c@{}}Sub-\\Specific \\ Avg. $\bar{\rho}$ \end{tabular}} &
        \textbf{\begin{tabular}[c]{@{}c@{}}Pooled\\ (h-avg) \\ Avg. $\bar{\rho}$ \end{tabular}} &
        \textbf{\begin{tabular}[c]{@{}c@{}}Pooled\\ (h-constant) \\ Avg. $\bar{\rho}$ \end{tabular}} &
        \textbf{\begin{tabular}[c]{@{}c@{}}Pooled (h-avg) \\ $>$ Sub-Specific \end{tabular}} &
        \textbf{\begin{tabular}[c]{@{}c@{}}Pooled (h-constant) \\ $>$ Sub-Specific \end{tabular}} \\ \midrule
        1 & 0.420 & 0.404 & 0.404 & 14.1$\%$ & 12.7$\%$ \\
        2 & 0.449 & 0.428 & 0.427 & 12.0$\%$  & 13.4$\%$ \\
        3 & 0.374 & 0.360 & 0.360 & 27.9$\%$  & 30.9$\%$ \\
        4 & 0.361 & 0.350 & 0.350 & 23.5$\%$  & 26.7$\%$ \\
        5 & {0.477} & {0.457} & {0.457} & 11.0$\%$  & 11.9$\%$ \\
        6 & 0.340 & 0.331 & 0.329 & 29.1$\%$  & 27.9$\%$ \\
        7 & 0.361 & 0.341 & 0.342 & 24.3$\%$  & 29.4$\%$ \\
        8 & 0.279 & 0.274 & 0.276 & {39.3$\%$}  & {42.5$\%$} \\ \midrule
        All & 0.382 & 0.368 & 0.368 & 22.6$\%$ & 24.4$\%$ \\ \bottomrule
    \end{tabular}
    \caption{Pooled versus subject-specific results for higher visual ROIs. For each subject, the average test set correlation $\rho$ as described in Section~\ref{sec: accuracy metric} across all higher visual ROIs is reported for CL-tuned AlexNet (Section~\ref{sec: contrastive learning model}) and pooled CL-tuned models using both a embedding layer dimension equal to the average voxel dimension of all subjects (h-avg) and an embedding layer dimension equal to the maximum voxel dimension over all subjects and ROIs (h-constant) (Section~\ref{sec: pooled models}). The percentage of voxels for which each version of pooled model was more predictive on average across the test set compared to the subject specific model is also shown. } 
    \label{table:pooled_table_higher}
\end{table}

\section{Downstream Classification Performance of ROI-tuned Models} \label{sec: downstream appendix}
\begin{figure}
    \centering
    \includegraphics[scale=1.15]{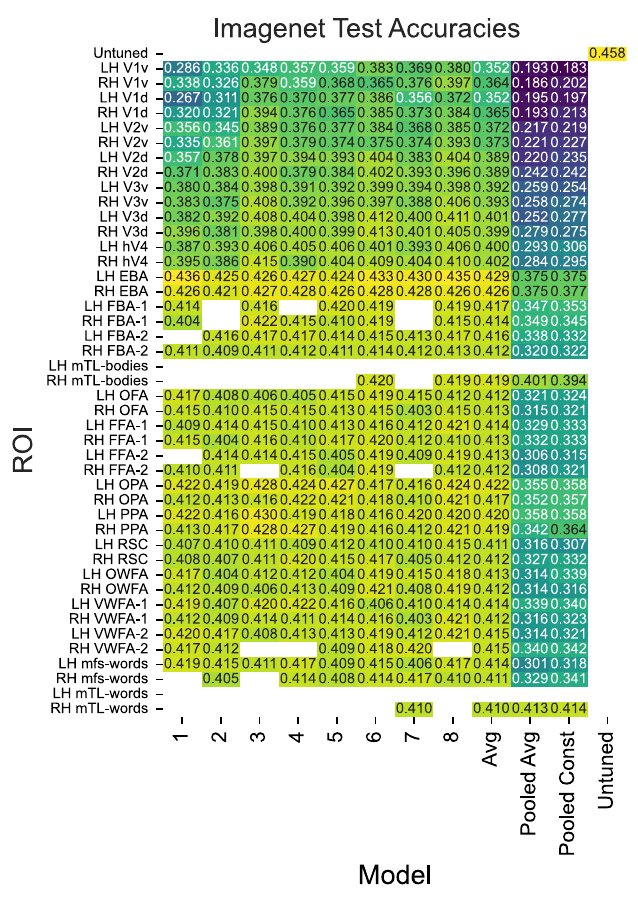}
    \caption{Test set accuracies for Caltech256 dataset for models for each ROI (both subject-specific and pooled), as well as untuned AlexNet for comparison.}
    \label{fig:imagenet_results_mat}
\end{figure}

\begin{figure}
    \centering
    \includegraphics[scale=1.15]{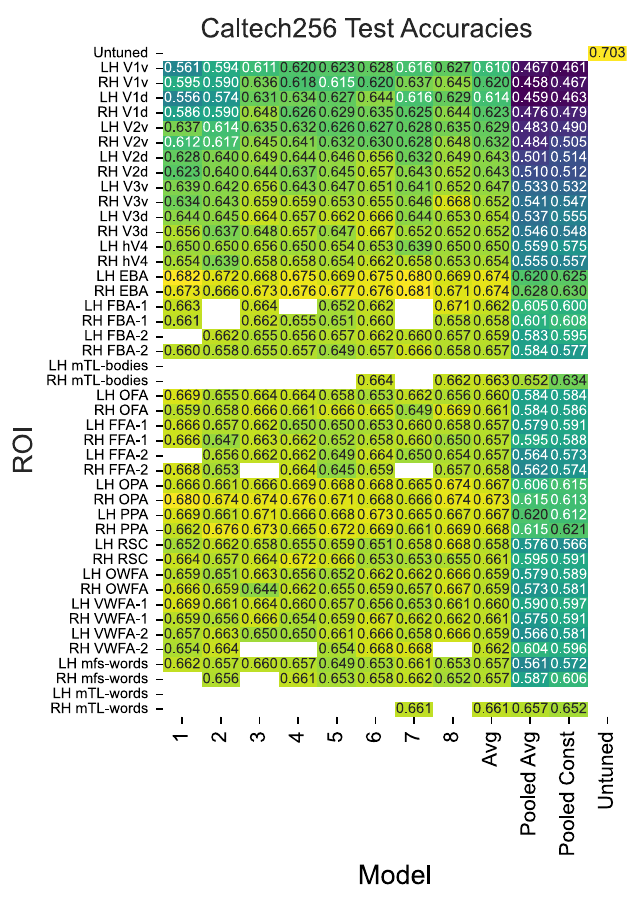}
    \caption{Test set accuracies for Places365 dataset for models for each ROI (both subject-specific and pooled), as well as untuned AlexNet for comparison.}
    \label{fig:caltech256_results_mat}
\end{figure}

\begin{figure}
    \centering
    \includegraphics[scale=1.15]{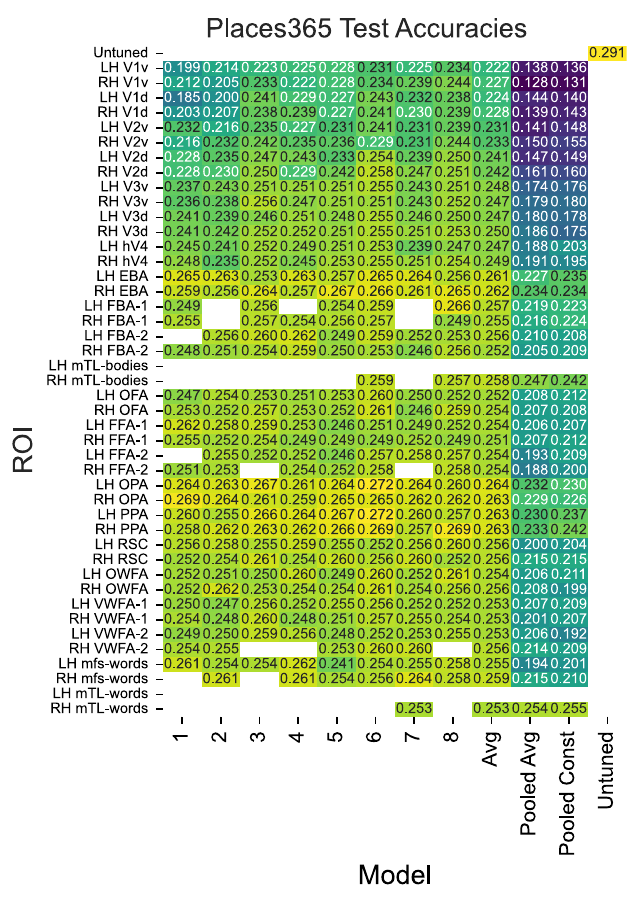}
    \caption{Test set accuracies for ImageNet dataset for models for each ROI (both subject-specific and pooled), as well as untuned AlexNet for comparison.}
    \label{fig:places365_results_mat}
\end{figure}

In Figures \ref{fig:imagenet_results_mat}, \ref{fig:caltech256_results_mat}, and \ref{fig:places365_results_mat}, we visualize the test set accuracy obtained for all three datasets, with both the subject-specific and pooled models for each ROI. 

\newpage

\subsection{Model Landscapes} \label{sec: model landscapes appendix}

\begin{figure}[htbp]
    \centering
    \includegraphics[width=\linewidth]{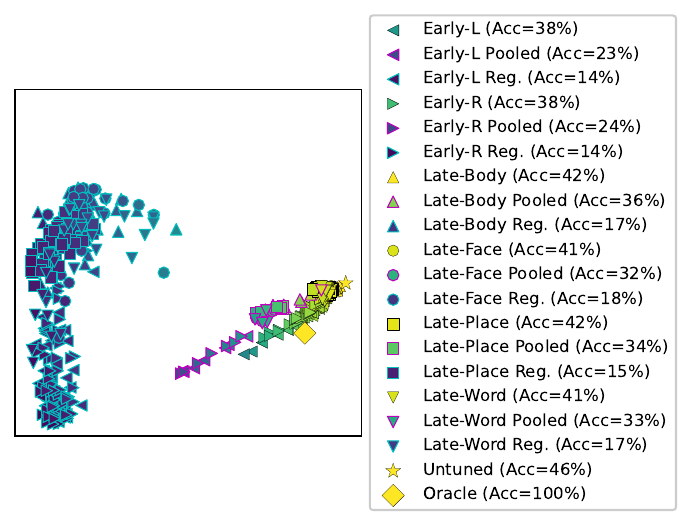}
    \caption{Model landscape using the multidimensional scaling embedding of the Bhattacharya-based dissimilarity for the oracle predictions (large diamond marker), untuned AlexNet (star marker), subject-specific ROI-specific CL fine-tuned models (black borders), subject-pooled ROI-specific fine-tuned models (magenta borders), and subject-specific ROI-specific regression fine-tuned models (cyan borders) . Marker face colors indicate classification accuracy clipped to fine-tuned range. Average classification per ROI group and model tuning appears in the legend.   }
    \label{fig:overall_landscape}
\end{figure}

Figure~\ref{fig:overall_landscape} shows the landscape of the full set of fine-tuned models along with the untuned AlexNet and an oracle's predictions based on true labels.    The regression tuned models have very low classification and distinct from the other models. The pooled models are distinct from subject-specific (except for a couple of ROIs which are present in only a single subject). 

\begin{figure}[H]
    \centering
    \includegraphics[width=0.57\linewidth]{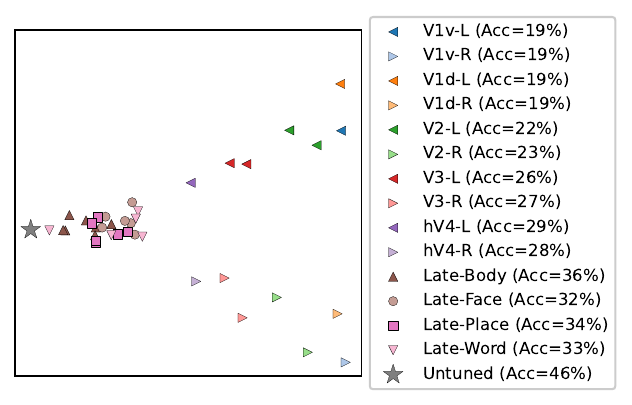}
    \includegraphics[width=0.4\linewidth]{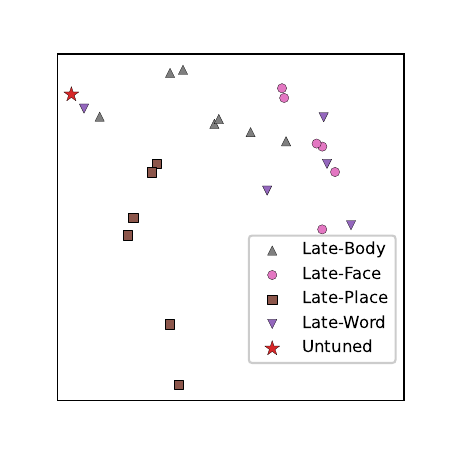}
    \caption{Model landscapes for untuned AlexNet (star marker) and pooled-subject ROI-specific ROIs. Left plot is all ROIs and right plot is  higher level (late).}
    \label{fig:pooled_landscape_key}
\end{figure}

Figure~\ref{fig:pooled_landscape_key} shows the landscapes for pooled models fine-tuned with CL.

\begin{figure}[H]
    \centering
        \includegraphics[width=0.5\linewidth]{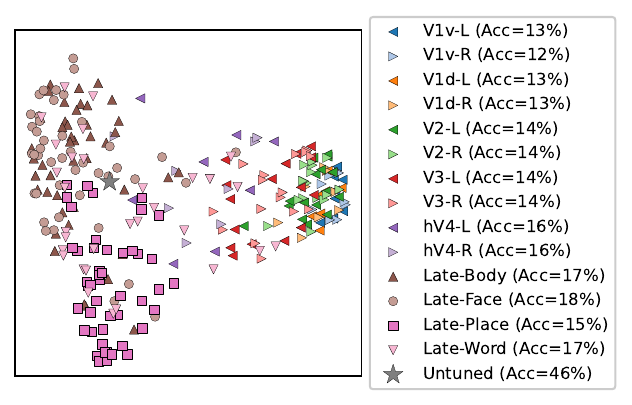}    
   \includegraphics[width=0.45\linewidth]{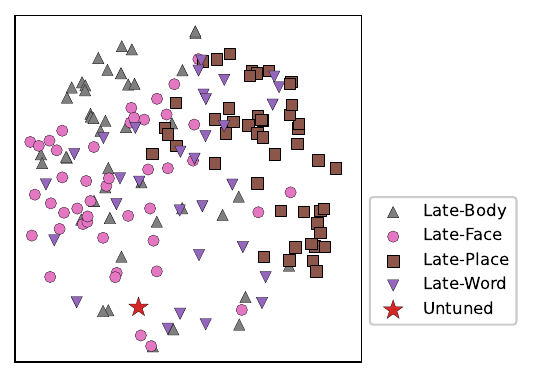}
    \caption{Model landscape for subject-specific and ROI-specific regression fine-tuned models and  AlexNet (star marker). Left plot is all ROIs and right plot is higher level (late). }
    \label{fig:reg_landscape}
\end{figure}

Figure~\ref{fig:reg_landscape} shows the landscapes for models fine-tuned with regression. 

\end{appendix}

\clearpage

\bibliography{ref}

\end{document}